\documentclass[letterpaper, 10 pt, conference]{ieeeconf}  

\IEEEoverridecommandlockouts                              

\usepackage{mathrsfs}
\usepackage{helvet}         		
\usepackage{type1cm}        		
\usepackage{graphics} 		
\usepackage{graphicx}        
\usepackage{epsfig} 			
\usepackage{multicol}        		
\usepackage[bottom]{footmisc}	

\usepackage{amssymb}
\usepackage[ruled]{algorithm2e}
\usepackage{subfigure}
\usepackage{newtxtext}       %
\usepackage{newtxmath}       
\usepackage{caption}
\usepackage{gensymb}
\usepackage{booktabs}
\usepackage[table,xcdraw]{xcolor}
\usepackage{epstopdf}
\usepackage{url}
\usepackage[font={small}]{caption}
\usepackage{commath}

\DeclareMathAlphabet{\mathcal}{OMS}{cmmi}{m}{n}

\DeclareFontFamily{U}{mathc}{}
\DeclareFontShape{U}{mathc}{m}{it}%
{<->s*[1.03] mathc10}{}
\DeclareMathAlphabet{\mathcal}{U}{mathc}{m}{it}

\newcommand{\fbf}{\mathbf{f}}

\newcommand{\lbf}{\mathbf{l}}

\newcommand{\sbf}{\mathbf{s}}

\newcommand{\xbf}{\mathbf{x}}

\newcommand{\Ebf}{\mathbf{E}}

\newcommand{\Acal}{\mathcal{A}}
\newcommand{\Bcal}{\mathcal{B}}

\newcommand{\Dcal}{\mathcal{D}}

\newcommand{\Fcal}{\mathcal{F}}
\newcommand{\Gcal}{\mathcal{G}}

\newcommand{\Lcal}{\mathcal{L}}

\newcommand{\Rcal}{\mathcal{R}}

\newcommand{\Ucal}{\mathcal{U}}

\newcommand{\Xcal}{\mathcal{X}}

\newcommand{\Pscr}{\mathscr{P}}

\newcommand{\Fsf}{\mathsf{F}}
\newcommand{\Gsf}{\mathsf{G}}

\newcommand{\thetabf}{\boldsymbol{\theta}}

\DeclareMathOperator*{\argmin}{arg\,min}

\title{\LARGE \bf
Autonomous Exploration Under Uncertainty via \\ Deep Reinforcement Learning on Graphs
}

\author{Fanfei Chen, John D. Martin, Yewei Huang, Jinkun Wang and Brendan Englot
\thanks{F. Chen, J. D. Martin, Y. Huang, J. Wang and B. Englot are with the Department of Mechanical Engineering, Stevens Institute of Technology, Castle Point on Hudson, Hoboken, NJ, 07030. 
        {\tt\small \{fchen7, jmarti3, yhuang85, jwang92, benglot\}@stevens.edu}}
}

\begin{document}

\maketitle
\thispagestyle{empty}
\pagestyle{empty}

\begin{abstract}

We consider an autonomous exploration problem in which a range-sensing mobile robot is tasked with accurately mapping the landmarks in an \textit{a priori} unknown environment efficiently in real-time; it must choose sensing actions that both curb localization uncertainty and achieve information gain. For this problem, belief space planning methods that forward-simulate robot sensing and estimation may often fail in real-time implementation, scaling poorly with increasing size of the state, belief and action spaces. We propose a novel approach that uses graph neural networks (GNNs) in conjunction with deep reinforcement learning (DRL), enabling decision-making over graphs containing exploration information to predict a robot's optimal sensing action in belief space. The policy, which is trained in different random environments without human intervention, offers a real-time, scalable decision-making process 
whose high-performance exploratory sensing actions yield accurate maps and high rates of information gain.

\end{abstract}
\label{sec:1}
\section{Introduction}

It is challenging to solve an autonomous mobile robot exploration problem \cite{Thrun2005} in an \textit{a priori} unknown environment when localization uncertainty is a factor, which may compromise the accuracy of the resulting map. Due to our limited ability to plan ahead in an unknown environment, a key approach to solving the problem is often to estimate and select the immediate next best viewpoint. The next-view candidates may be generated by enumerating frontier locations or using sampling-based methods to plan beyond frontiers. A utility function may be used to evaluate the next-view candidates and select the optimal candidate. It is common to forward-simulate the actions and measurements of each candidate and choose the best one as the next-view position \cite{Julian2014, Charrow2015CSQMI}. However, as the size of the robot's state and action space increases, the computation time of this decision-making process grows substantially. Although learning-based exploration algorithms can accurately predict optimal next-view positions with nearly constant computation time, these algorithms may suffer when transferring a learned policy to a new unknown environment due to poor generalizability. Our recent prior work \cite{Chen2019ISRR} proposed a generalized graph representation for mobile robot exploration under localization uncertainty, compatible with graph neural networks (GNNs). However, the hyperparameters are difficult to tune and the performance in test is unstable because the size of the graph is always changing throughout the exploration process.

\begin{figure}[t]
\centering
\includegraphics[width=0.90\columnwidth]{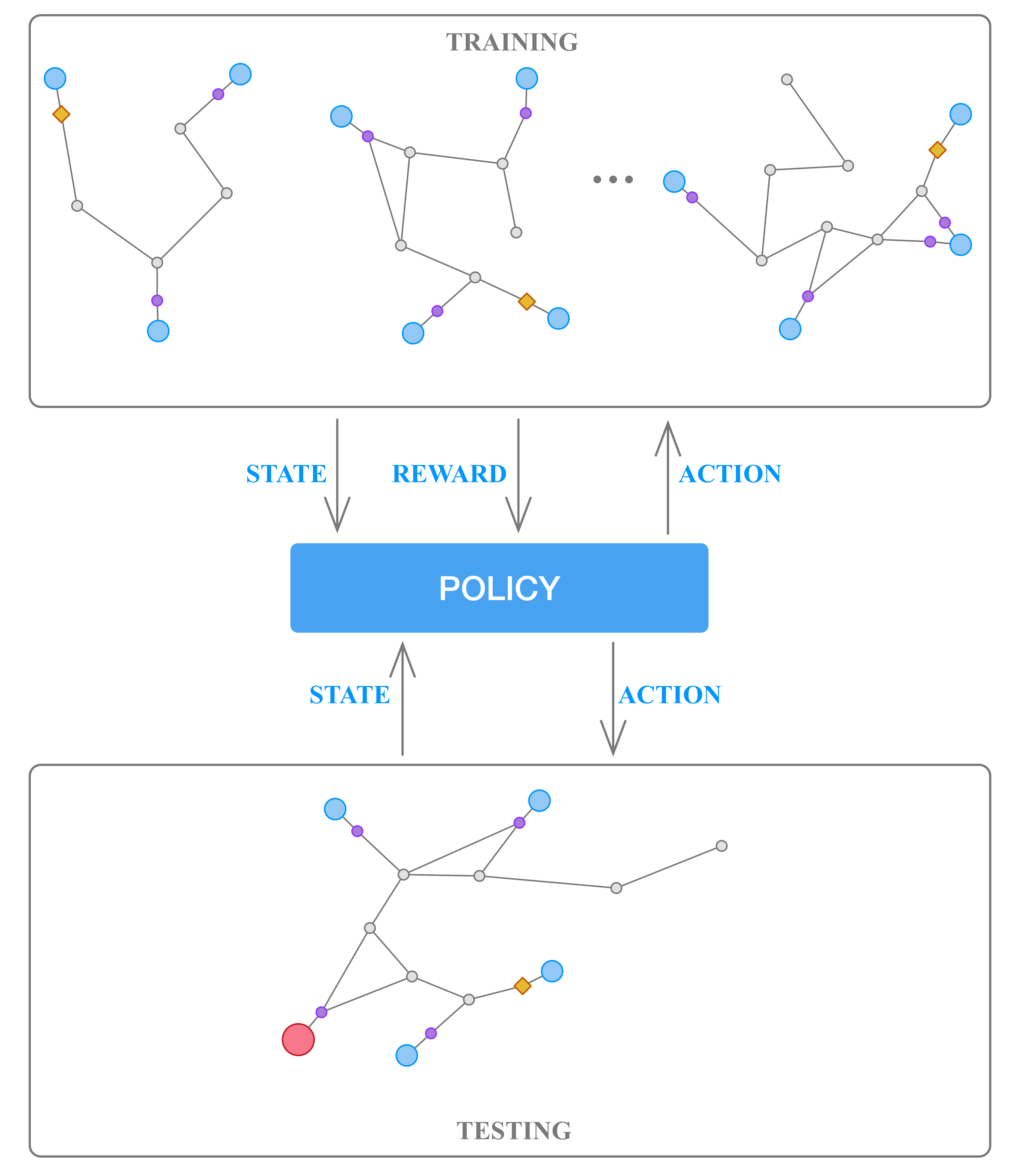}
\caption{\textbf{An illustration of our framework.} In the \textit{training phase}, the robot is trained in different random environments with a random initial robot location and distribution of landmarks. The current state is represented by an \textit{exploration graph}. The robot pose history, observed landmarks, the current pose and candidate frontier waypoints are indicated by gray circles, purple circles, an orange square and blue circles respectively. The parameters are optimized in the policy network and/or the value network according to the reward given by the environment. Then in the \textit{testing phase}, we generate a different set of random environments. The optimal action is estimated by the trained policy, which is indicated in red.}
\label{rl}
\vspace{-5mm}
\end{figure}

In this paper, we aim to learn a more robust and generalizable policy for exploration using deep reinforcement learning (DRL), where the optimal view can be estimated by the DRL policy instead of explicitly computing the expected utility of each candidate sensor view. Hence our method achieves a nearly constant-time, real-time viable decision-making process. Our \textit{exploration graph} is a generalized state representation provided as input. Instead of learning from camera images or metric maps, such a graph reduces the dimensionality of our state space, so the policy can be exposed to and tested in a wide diversity of environments. Fig. \ref{rl} summarizes our general approach. The policy is trained in environments with randomly distributed landmarks, and while exploring, predictions of the next best view are provided by the learned policy. The policy is generalizable to environments of different sizes, and with different quantities of landmarks, whose graph topologies are well-represented despite our framework being trained in simpler environments. 

In this paper, we present the first framework to perform reinforcement learning with graph neural networks for the purpose of teaching a mobile robot how to efficiently explore unknown environments under localization uncertainty, while building an accurate map. We demonstrate that this allows our \textit{exploration graph} abstraction to support robust and generalizable learning, which also scales favorably in problem size and performance against state-of-the-art approaches. 


\subsection{Related Work}
Information-theoretic exploration methods often use a prediction of the mutual information (MI) between a robot's future sensor observations and the cells of an occupancy map to evaluate the next-view candidates being considered \cite{Julian2014, Charrow2015CSQMI}. A similar approach is applied specifically to the frontiers of Gaussian process occupancy maps \cite{Jadidi2018} to efficiently predict the most informative sensing action. However, robot localization uncertainty is not considered in these approaches.

Active simultaneous localization and mapping (SLAM) exploration methods consider both map entropy and localization and/or map uncertainty. The active pose SLAM approach of \cite{Valencia2012} uses the sum of map and robot trajectory entropy to guide a robot's exploration of an unknown environment. Particle filtering is employed in \cite{Stachniss2005} to manage uncertainty during exploration by transforming trajectory uncertainty into particle weights. \textit{Virtual landmarks} are proposed in the Expectation-Maximization (EM) exploration algorithm \cite{Wang2017} to model the impact of a robot's pose uncertainty on the accuracy of an occupancy map, employing a utility function favoring sensing actions that drive down the collective uncertainty of the virtual landmarks residing in all map cells.

The above methods use forward simulation to evaluate the mapping outcomes and uncertainty of next-view candidates, and so their computational cost increases substantially with an increasing number of next-view candidates. Learning-based exploration methods offer the prospect of improved scalability for selecting the next best view, with and without localization uncertainty. Learning-aided information-theoretic approaches \cite{Bai2015}, \cite{Bai2016} are proposed to reduce the cost of predicting MI. Deep neural networks in \cite{Bai2017}, \cite{Chen2019} and \cite{Niroui2019} can be trained to predict the optimal next-view candidate through supervised learning and  reinforcement learning, by taking occupancy grid maps as input data. However, the state space of such maps is very large, and a learned policy may fail when tested in a completely new environment. 

Graphs can offer generalized topological representations of robot navigation, permitting a much smaller state space than metric maps. GNNs incorporate graphs into neural network models, permitting a variety of queries to be posed over them \cite{Scarselli2009}. Sanchez-Gonzalez et al. \cite{Sanchez2018} proposed to use graph models of dynamical systems to solve control problems with Graph Nets \cite{Battaglia2018}. Exploration under localization uncertainty is achieved in our prior work \cite{Chen2019ISRR} through supervised learning over graphs, which gives an unstable result because the number of nodes in each graph differs, and the hyperparameters of the loss function are hard to tune. A novel graph localization network is proposed in \cite{KChen2019} to guide a robot through visual navigation tasks. Wang et al. \cite{Wang2018} introduces GNNs as policy networks and value networks, through their integration into DRL, to solve control problems.

We combine the exploration graphs proposed in \cite{Chen2019ISRR} with DRL algorithms to learn robot exploration policies. The policy can be trained without human intervention. Our proposed approach offers a generalized representation of a SLAM-dependent mobile robot's state and environment, and a self-learning mechanism offering wide applicability.

\subsection{Paper Organization}
The definition of our mobile robot active SLAM exploration problem, and the corresponding exploration graph abstraction, is provided in Section \ref{sec:Formulation}. In Section \ref{sec:Algorithms}, a framework for DRL with GNNs to address robot exploration under uncertainty is presented. Experimental results are given in Section \ref{sec:experiments}, with conclusions in Section \ref{sec:conclusions}.

\begin{figure}[t]
\centering
\subfigure[]{\includegraphics[width=0.8\columnwidth]{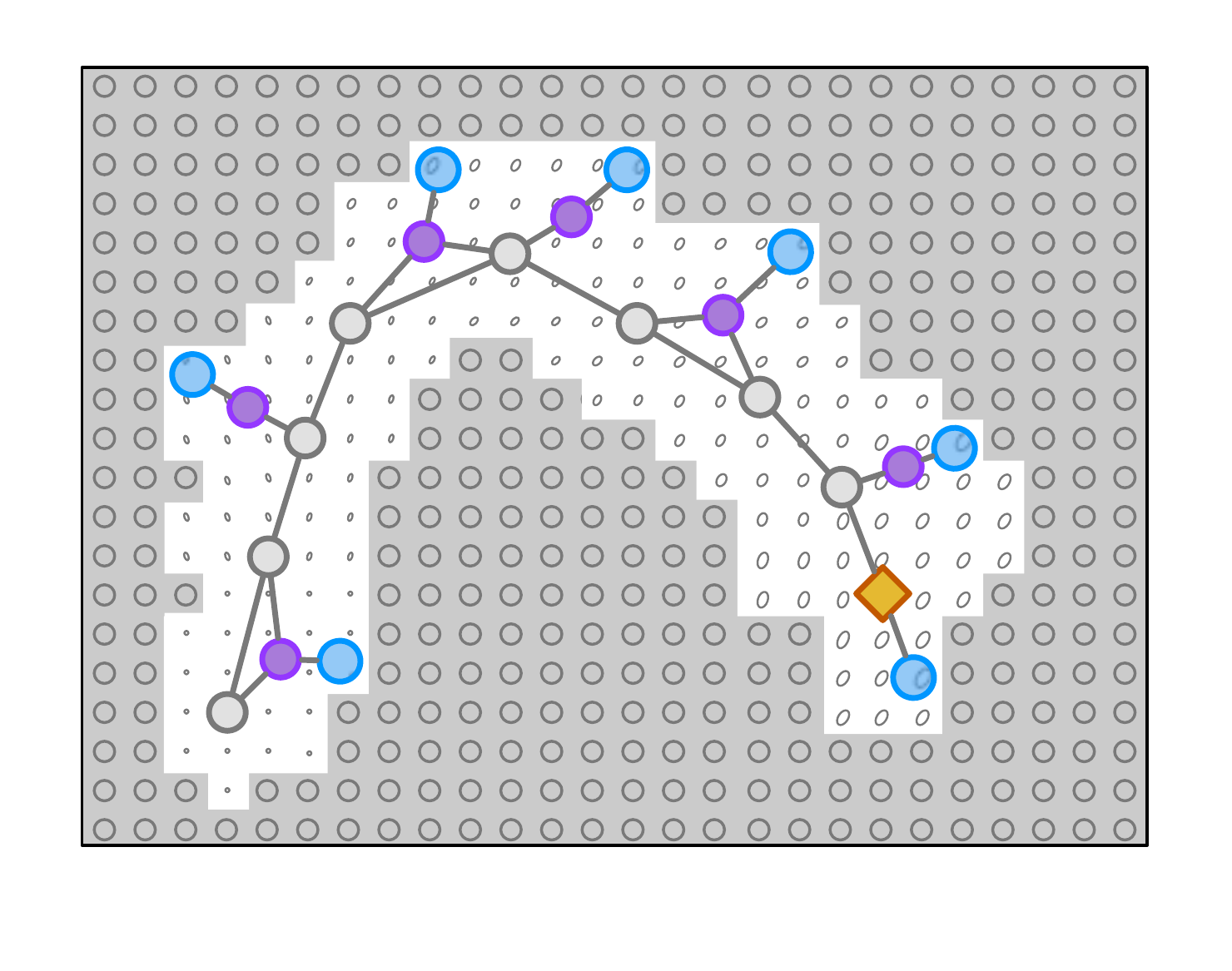}\label{eg1}}
\subfigure[]{\includegraphics[width=0.7\columnwidth]{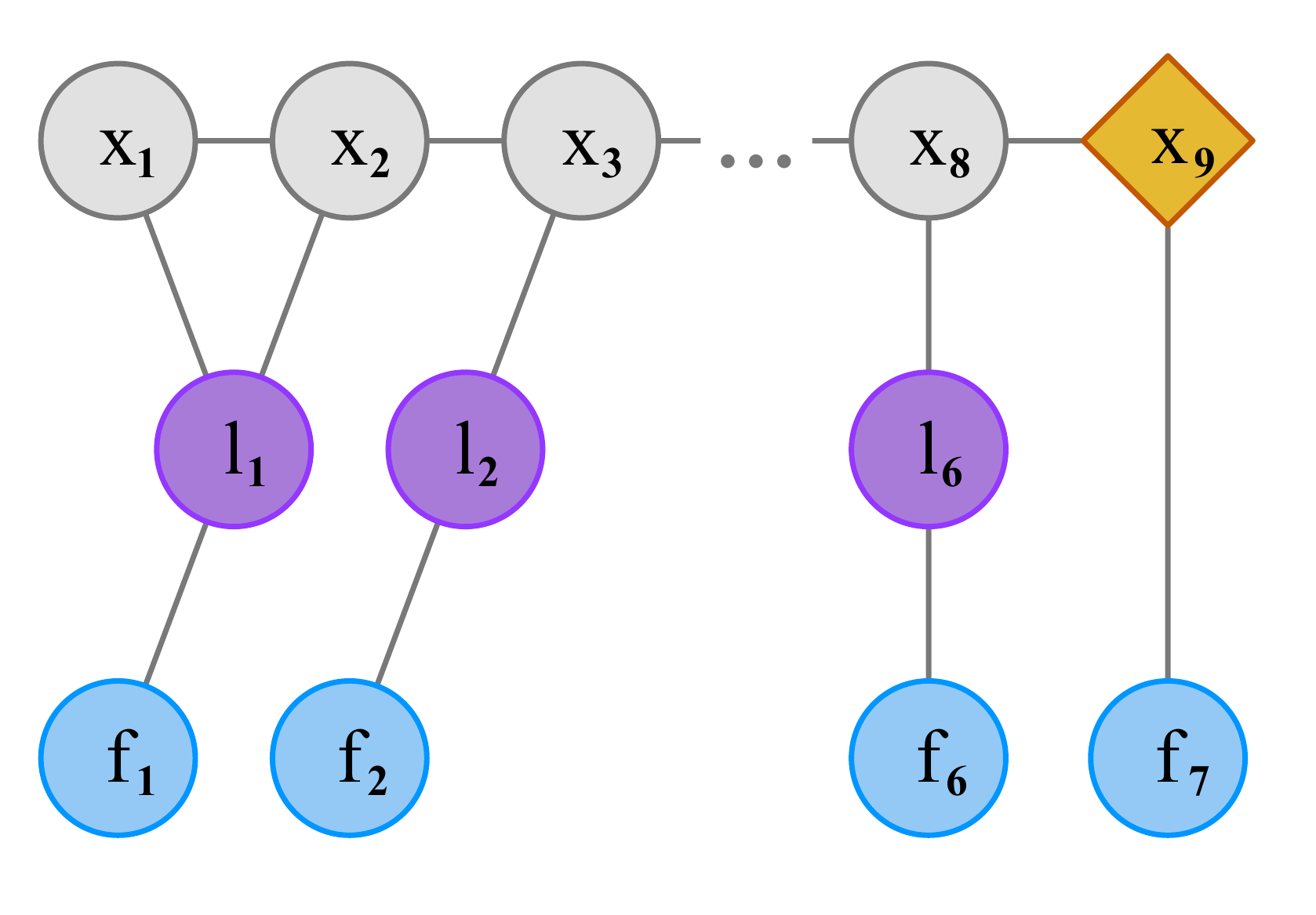}\label{eg2}}
\caption{\textbf{Formulating and extracting the exploration graph.} Top: the grayscale color represents the probability of occupancy of a given map cell (assumed to be 0.5 for unobserved cells). The ellipse in each cell represents the estimation error covariance of each \textit{virtual landmark}, with true landmarks shown in purple. Past robot poses, the current pose, and candidate frontiers are indicated by gray circles, an orange diamond, and blue circles respectively. Landmark nodes and the current pose node are connected with their nearest frontiers. Bottom: An input exploration graph is extracted from the current exploration state. Each edge in this graph is weighted with the Euclidean distance between the two vertices connected.}
\label{exploration_graph}
\vspace{-5mm}
\end{figure}

\section{Problem Formulation and Approach}
\label{sec:Formulation}
\subsection{Simultaneous Localization and Mapping Framework}
We adopt a graph-based approach in the SLAM framework supporting our robot's exploration. We then solve the SLAM problem as a least-squares smoothing problem. The robot motion model and measurement models are defined as:
\begin{align}
& \mathbf x_i = h_i(\mathbf x_{i-1}, \mathbf u_i) + \mathbf w_i,\quad \mathbf w_i \sim \mathcal N(\mathbf 0, Q_i),\\
& \mathbf z_{k} = g_{k}(\mathbf x_{i_k}, \mathbf l_{j_k}) + \mathbf v_{k},\quad \mathbf v_{k} \sim \mathcal N(\mathbf 0, R_{k}),
\end{align}
where $\mathcal X = \{\mathbf x_i\}_{i=1}^t$ are poses along the trajectory, $\mathcal L = \{\mathbf l_j\}_{j=1}^m$ are $m\in\mathbb{N}$ landmarks, and $\mathcal{U}=\{ \mathbf{u_i} \}_{i=1}^t$ is a given sequence of low-level actions. Then we solve the SLAM problem as a least-squares problem:
\begin{equation}
\begin{aligned}
	\mathcal X^*, \mathcal L^* = \argmin_{\mathcal X, \mathcal L} \sum_{i} \norm{\mathbf{x_i}-h_i(\mathbf{x_{i-1}}, \mathbf{u_i})}^2_{Q_i} \\ + \sum_{k} \norm{\mathbf{z_k}-g_k(\mathbf{x_{i_k}}, \mathbf{l_{j_k}})}^2_{R_{k}}.
	\end{aligned}
\end{equation}
$\mathcal X^*$ and $\mathcal L^*$ are obtained using the GTSAM \cite{gtsam} implementation of iSAM2 \cite{Kaess2012} to perform nonlinear least-squares smoothing over a factor graph. The Gaussian marginal distributions and Gaussian joint marginal distributions are produced from this graph-based inference procedure.

We next introduce the \textit{virtual map} of the EM algorithm for exploration under localization uncertainty \cite{Wang2017}, which is a uniformly discretized grid map comprised of \textit{virtual landmarks}, $\tilde{\lbf}_k \in \tilde{\Lcal}$, to represent the uncertainty and occupancy probability of every map cell. Each cell of the virtual map contains a virtual landmark with a large initial covariance (illustrated in Figures \ref{exploration_graph}(a) and \ref{sim_env}). We use A-optimality \cite{Kaess2009} for our uncertainty criterion:
\begin{equation}
\label{covariance}
    \phi_\text{A}(\Sigma) = \text{tr}(\Sigma),
\end{equation}
where $\Sigma$ is the error covariance matrix for all virtual landmarks in the map. Accordingly, the uncertainty of the current state is quantified by the trace of the covariance of all virtual landmarks. The definition of the utility function is as follows:
\begin{equation}
\label{utility_function}
U(\tilde{\Lcal}) = \sum_{\tilde{\lbf}_k \in \tilde{\Lcal}} \phi_\text{A} (\Sigma_{\tilde{\lbf}_k}).
\end{equation}

\subsection{Exploration Graph}

The definition of the \textit{exploration graph} is $\mathcal{G}=\mathcal{(V, E)}$, where $\mathcal{V} = \Xcal\cup \Lcal\cup \Fcal$, and where $\Xcal, \Lcal,$ and $\Fcal$ are sets of previous poses, landmarks, and candidate frontier nodes respectively. The edges $\mathcal E$ connecting pose to pose $\mathbf x_i\ \text{\textemdash}\ \mathbf x_{i+1}$, pose to landmark $\mathbf x_{i_k}\ \text{\textemdash}\ \mathbf l_{j_k}$, landmark to frontier $\mathbf l_{j_k} \text{\textemdash}\ \mathbf f_{n_k}$, and the current pose to its nearest frontier $\xbf_{t} \text{\textemdash}\ \mathbf f_{n_t}$ are weighted with the Euclidean distances between those nodes. 
Each node $\mathbf{n_i} \in \mathcal V$ has a feature vector
\begin{align}
& \sbf_i=[s_{i_1}, s_{i_2}, s_{i_3}, s_{i_4}, s_{i_5}],\nonumber\\
& s_{i_1}= \phi_\text{A}(\Sigma_i), \label{s1} \\
& s_{i_2}= \sqrt{{(x_i-x_t)}^2 + {(y_i-y_t)}^2}, \label{s2} \\
& s_{i_3}= \text{arctan2}(y_i-y_t,x_i-x_t), \label{s3} \\
& s_{i_4}= p(m_i=1), \label{s4} \\
& s_{i_5}= 
        \begin{cases}
            0 & \mathbf{n_{i}} = \mathbf{x_t} \\
            1 & \mathbf{n_{i}} \in \{ \mathbf{f_n} \} \\
            -1 & \text{otherwise}
        \end{cases}. \label{s5}
\end{align}

The contents of the feature vector are as follows. In Eq. (\ref{s1}), the A-Optimality criterion, derived from our virtual map $\tilde{\Lcal}$, is used to quantify a node's uncertainty. Eqs. (\ref{s2}) and (\ref{s3}) provide relative pose information; the Euclidean distance and the relative orientation between the current robot pose and each node in the exploration graph. Occupancy information is extracted from an occupancy grid map, where $m_i$ is the occupancy of the map cell associated with node $n_i$ in Eq. (\ref{s4}). Eq. (\ref{s5}) provides an indicator variable, denoting the current pose to be 0, all frontiers to be 1, and all other nodes -1.

We note that frontier nodes are sampled from the boundary cells between free and unexplored space in the occupancy map. The landmarks and the current robot pose are the only nodes connected to frontiers, and each connects only to its nearest frontier node. It is possible for multiple landmark nodes to be connected to the same frontier node, but not for a landmark or pose node to connect to multiple frontier nodes; only the nearest frontiers are connected into the graph. The composition of an exploration graph is shown in Fig. \ref{exploration_graph}.

\section{Algorithms}
\label{sec:Algorithms}
\subsection{Graph Neural Networks}

We explore three different graph neural networks in this paper: Graph Convolutional Networks (GCNs) \cite{Kipf2017}, Gated Graph Neural Networks (GG-NNs) \cite{Li2016} and Graph U-Nets (g-U-Nets) \cite{Gao2019}. The GCN model performs convolution operations on graphs using the information of neighbors around each node. GG-NNs adopt a gated sequential unit to update the node feature vectors of graphs. Finally, g-U-Nets include pooling and unpooling layers to achieve an encoder-decoder mechanism similar to U-Net \cite{Ronneberger2015}. Each GNN model has three hidden layers and a multilayer perceptron (MLP) output model. We add a dropout layer between each GNN layer to prevent overfitting. These GNNs are used in our RL framework as policy networks and/or value networks.

\subsection{Deep Reinforcement Learning}
Reinforcement learning describes a sequential decision making problem, whereby an agent learns to act optimally from rewards collected after taking actions. In our setting, we consider changes to the exploration graph that result from our selection of frontier nodes, which represent waypoints lying on the boundaries between mapped and unmapped areas. At each step $k\in\mathbb{N}$ the environment state is captured by the exploration graph $\Gcal_k\in\Gsf$ \footnote{The exploration graph $\Gcal$ encodes the full history of robot poses.}. The robot chooses a frontier node to visit based on the graph topology: $\fbf_k\in\Fsf_{\Gcal_k}$. This causes a transition to $\Gcal_{k+1}\in\Gsf$ and a scalar reward $R\in\mathbb{R}$ to be emitted. The interaction is modeled as a Markov Decision Process $\left<\Gsf, \Fsf, \textnormal{Pr}, \gamma \right>$ \cite{Puterman1994}, associated with the transition kernel $\textnormal{Pr} \colon \Gsf \times \Fsf \rightarrow \Pscr(\Gsf\times\mathbb{R})$ that defines a joint distribution over the reward and next exploration graph, given the current graph and frontier node pair. Here, $\gamma \in [0,1)$ is a discount factor used to control the relative importance of future rewards in the agent's learning objective.

\begin{algorithm}[t]
\caption{Reward Function}
\label{alg:reward}
\textbf{input:} Exploration graph $\Gcal$, Frontier node $\fbf$\\
{\color{gray} \# Normalize the raw reward Alg. \ref{alg:raw_reward}}\\
$\Rcal_{\Gcal} = \{ r_{\fbf'} = \texttt{raw\_reward}(\Gcal,\fbf') \ \forall \ \fbf' \in \Acal_{\Gcal}\}$\\
$l = \min\Rcal_{\Gcal}$, $u = \max\Rcal_{\Gcal}$\\
$r_{\fbf} \gets (r_{\fbf} - l)/(u-l)$\\
{\color{gray} \# Compute projection based on nearest frontier}\\
$\fbf_t=\texttt{nearest\_frontier}(\xbf_t)$\\
\If{$u$ = \texttt{raw\_reward}$(\fbf_t)$}{
    \textbf{return} $r_{\fbf}-1$ {\color{gray} \# $r(\Gcal,\fbf)\in[-1,0]$}
}
\textbf{return} $2r_{\fbf}-1$ {\color{gray} \# $r(\Gcal,\fbf)\in[-1,1]$}
\end{algorithm}

\begin{algorithm}[t]
\caption{Raw reward}
\label{alg:raw_reward}
\textbf{input:} Exploration graph $\Gcal$, Frontier node $\fbf$\\
$\Ucal\gets$\texttt{path\_planner}$(\Gcal, \fbf)$\\
 {\color{gray} \# Compute raw reward with \eqref{utility_function}, Alg \ref{alg:utility} and cost-to-go}\\
\textbf{return} $U(\tilde{\Lcal}) - \texttt{compute\_utility}_{\tilde{\Lcal}}(\Ucal) - \alpha C(\Ucal)$
\end{algorithm}

\begin{algorithm}[t]
\caption{Compute Utility}
\label{alg:utility}
\textbf{global:} Virtual landmarks $\tilde{\Lcal}$\\
\textbf{input:} Sequence of low-level actions $\Ucal$\\
{\color{gray}\# Execute low-level actions with \cite{Wang2017} }
 $\tilde{\Lcal}\gets$\texttt{update\_virtual\_map}$(\Ucal)$\\
{\color{gray}\# Compute uncertainty estimate with \eqref{utility_function}}\\
\textbf{return} $U(\tilde{\Lcal})$\\
\end{algorithm}

\begin{algorithm}[t]
\caption{Exploration Training with RL}
\label{alg:train}
\textbf{initialize:} $\Gcal$, $\thetabf$, $step$\\
$\Dcal \gets \emptyset $\\
\While{\texttt{step} < \texttt{max\_training\_steps} }
{
\uIf{RL="DQN"}{
{\color{gray}\# Gather experience visiting frontier nodes}\\
$\fbf \gets \texttt{action\_sampling}(\mathcal{G})$\\
$\mathcal{G}', r \gets \texttt{visit\_frontier}(\Gcal, \fbf)$\\
$step \gets step+1$\\
$\Dcal \gets \Dcal \cup\{\Gcal, \fbf, r, \Gcal'\}$\\
$\mathcal{G} \gets \mathcal{G}'$\\
{\color{gray}\# Train policy with DQN algorithm}\\
$\pi_{\thetabf} \gets \texttt{dqn}(\Dcal)$\\
}
\uElseIf{RL="A2C"}{
{\color{gray}\# Gather experience visiting frontier nodes}\\
$\fbf \gets \pi_{\thetabf}(\mathcal{G})$\\
$\mathcal{G}', r \gets \texttt{visit\_frontier}(\Gcal, \fbf)$\\
$step \gets step+1$\\
$\Dcal \gets \Dcal \cup\{\Gcal, \fbf, r, \Gcal'\}$\\
$\mathcal{G} \gets \mathcal{G}'$\\
{\color{gray}\# Train policy with A2C algorithm}\\
\uIf{\texttt{$step \mod policy\_update\_steps=0$}}{
$\pi_{\thetabf} \gets \texttt{a2c}(\Dcal)$\\
$\Dcal \gets \emptyset $\\
}
}}
\textbf{return:} $\pi_{\thetabf}$
\end{algorithm}

In this paper we consider value-based methods and policy-based methods for model-free control. Value methods strive to maximize the expected sum of future rewards, or the \textit{value}:
\begin{align}
    Q_{\thetabf}(\Gcal,\fbf) &= \Ebf\left[\sum_{k=0}^\infty\gamma^k R(\Gcal_k,\fbf_k) \ \biggl| \ \Gcal_0=\Gcal,\fbf_0=\fbf\right].
\end{align}
For large-scale problems, the value function is represented with a parameterized function, such as a convolutional neural network or, in our study, a graph neural network. 

To train $Q_{\thetabf}$, we gather transition samples of the form $(\Gcal,\fbf,r,\Gcal')$ using an action sampling strategy that chooses the most uncertain action, which has the largest value at the output. According to \cite{Gal2015}, the uncertainty of the actions can be represented by the output of the dropout layer. During training, we first dropout $90\%$ of our data before the MLP output model, and as the training phase runs, we gradually decrease the dropout rate until it reaches $0\%$, which means the policy provides greedy action selection.
The parameters $\thetabf$ are adjusted using stochastic gradient descent to minimize the loss function $L \colon \Bcal \rightarrow \mathbb{R}$ over sampled minibatches $\Bcal \sim \Dcal = \{(\Gcal,\fbf,r,\Gcal')_k\}_{k\in\mathbb{N}}$. In this paper we consider the DQN \cite{Mnih2015} loss function:
\begin{equation}
\label{dqn_loss}
L_{\textnormal{DQN}}(\Dcal) = \Ebf_{\Bcal\sim\Dcal}[(r + \gamma \max_{\fbf' \in \Fsf_{\Gcal'}} Q(\Gcal', \fbf')-Q(\Gcal,\fbf))^2],
\end{equation}
which encodes the expected squared TD-error of one-step predictions over $\Bcal$. 

We also consider the policy-based method A2C \cite{Vinyals2017}, that directly trains a parameterized policy $\pi_{\thetabf} \colon \Gsf \rightarrow \Pscr(\Fsf)$ from data gathered following that policy. We use two separate GNN models to serve as the policy network and the value network, and train it with the loss function:
\begin{align}
\label{a2c_loss}
L_{\textnormal{A2C}}(\Dcal)&=\Ebf_{\Bcal\sim\Dcal}[L_{\textnormal{A2C}}^{(1)}+\eta L_{\textnormal{A2C}}^{(2)}],\\
L_{\textnormal{A2C}}^{(1)} &= [A(\Gcal,\fbf)-V(\Gcal))\log\pi(\fbf|\Gcal)+\beta(A(\Gcal,\fbf)]^2 \nonumber, \\ 
L_{\textnormal{A2C}}^{(2)} &= \sum_{\fbf \in \Fsf_{\Gcal}}\pi(\fbf|\Gcal)\log\pi(\fbf|\Gcal)\nonumber.
\end{align}
Here, $L_{\textnormal{A2C}}^{(1)}$ and $L_{\textnormal{A2C}}^{(2)}$ denote the loss terms for a single transition sample. The function $A(\Gcal,\fbf) = Q(\Gcal,\fbf)-V(\Gcal)$ is called the \textit{advantage}; it computes the difference between the state-action value function $Q$ and the state value function $V(\Gcal) = \max_{\fbf\in\Fsf_{\Gcal}}Q(\Gcal,\fbf)$. We use $\beta\in\mathbb{R}$ as a coefficient for the value loss, and we use $\eta\in\mathbb{R}^+$ as a coefficient for the entropy of the output, to encourage exploration within the RL solution space.

\subsection{Reward Function}

In Algorithm \ref{alg:reward}, we use linear normalization functions to map the range of the raw reward. If the optimal frontier is the one associated with the current pose, the maximum of the reward is 0, otherwise, it is 1 in all other cases. Algorithm \ref{alg:raw_reward} is designed based on the utility function of the EM exploration algorithm, Eq. (\ref{utility_function}), with an additional term penalizing the travel distance associated with an exploratory action. The raw reward is the difference in utility between the current state and the subsequent state by taking the actions $\Ucal$. The cost-to-go $C(\Ucal)$ with factor $\alpha$ encourages short travel distances in the raw reward function. The goal is to minimize this weighted combination of map uncertainty and travel distance with every given visit to a frontier node.

\subsection{Exploration Training with RL}
As shown in Algorithm \ref{alg:train}, using an action sampling strategy, the robot chooses a next-view frontier based on an \textit{exploration graph}. A reward and a new exploration graph are assigned after reaching the frontier most recently selected. The policy is trained using recorded experience data.

\vspace{2mm}

\section{Experiments and Results}
\label{sec:experiments} 

\begin{figure}[t]
\centering
\subfigure[An illustration of the simulation environment]{\includegraphics[width=0.99\columnwidth]{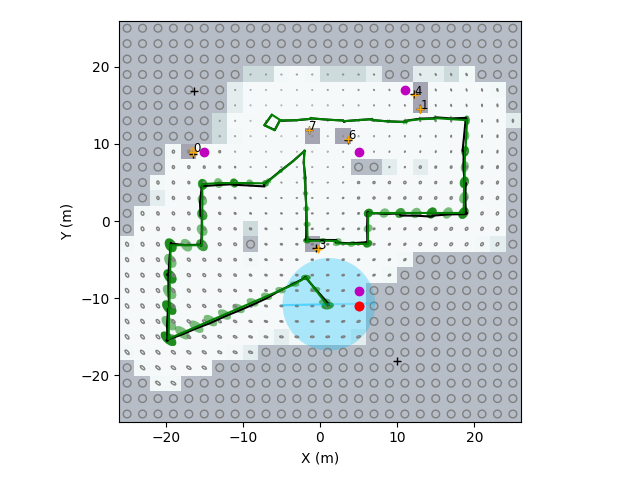}\label{sim_env}}
\subfigure[Average reward during the training process]{\includegraphics[width=0.97\columnwidth]{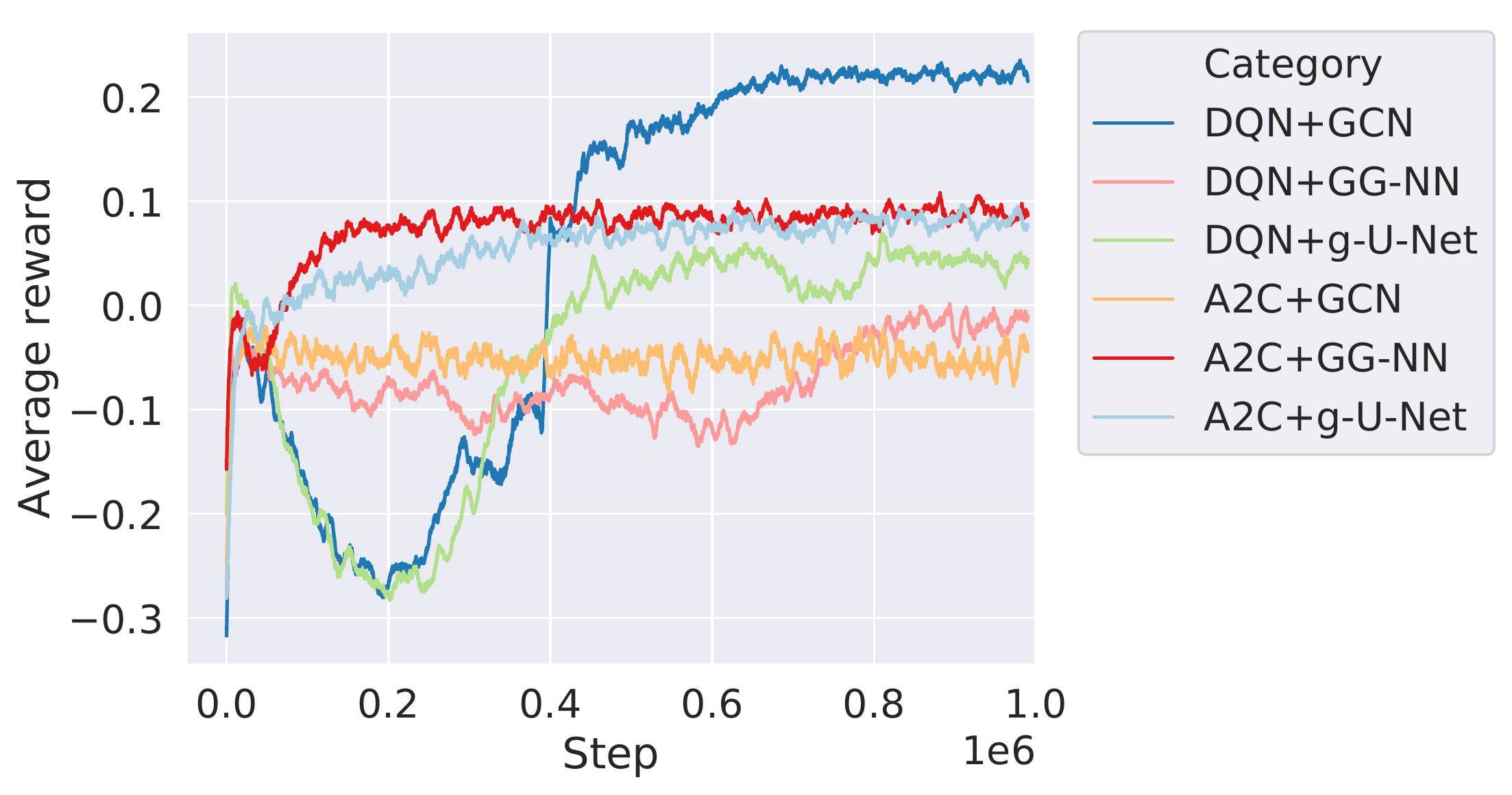}\label{training_ave_reward}}
\caption{The training performance of different methods on randomly generated simulation environments.}
\vspace{-0mm}
\end{figure}

\begin{figure}[t]
\centering
\includegraphics[width=0.9\columnwidth]{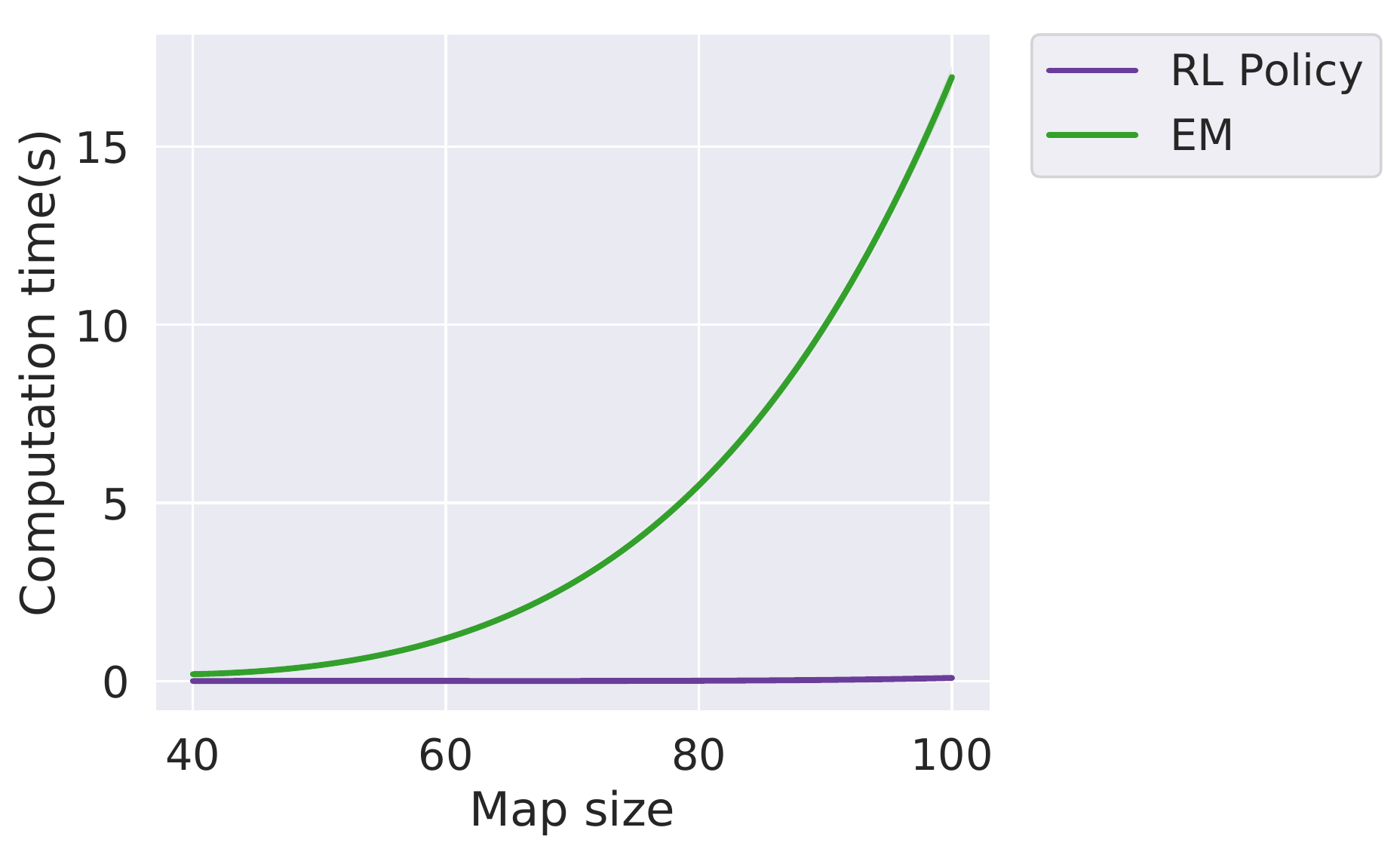}
\caption{Computation time for exploration decision-making on different map sizes, which are square in the dimension indicated. Timing was evaluated on a computer equipped with an Intel i9 8-core 3.6Ghz CPU and an Nvidia GeForce Titan RTX GPU. The average computation time is 0.04427s for the RL policy.}
\label{com_time}
\vspace{-0mm}
\end{figure}

\begin{figure*}[ht]
\centering
\subfigure[Average uncertainty of landmarks]{\includegraphics[height=36mm]{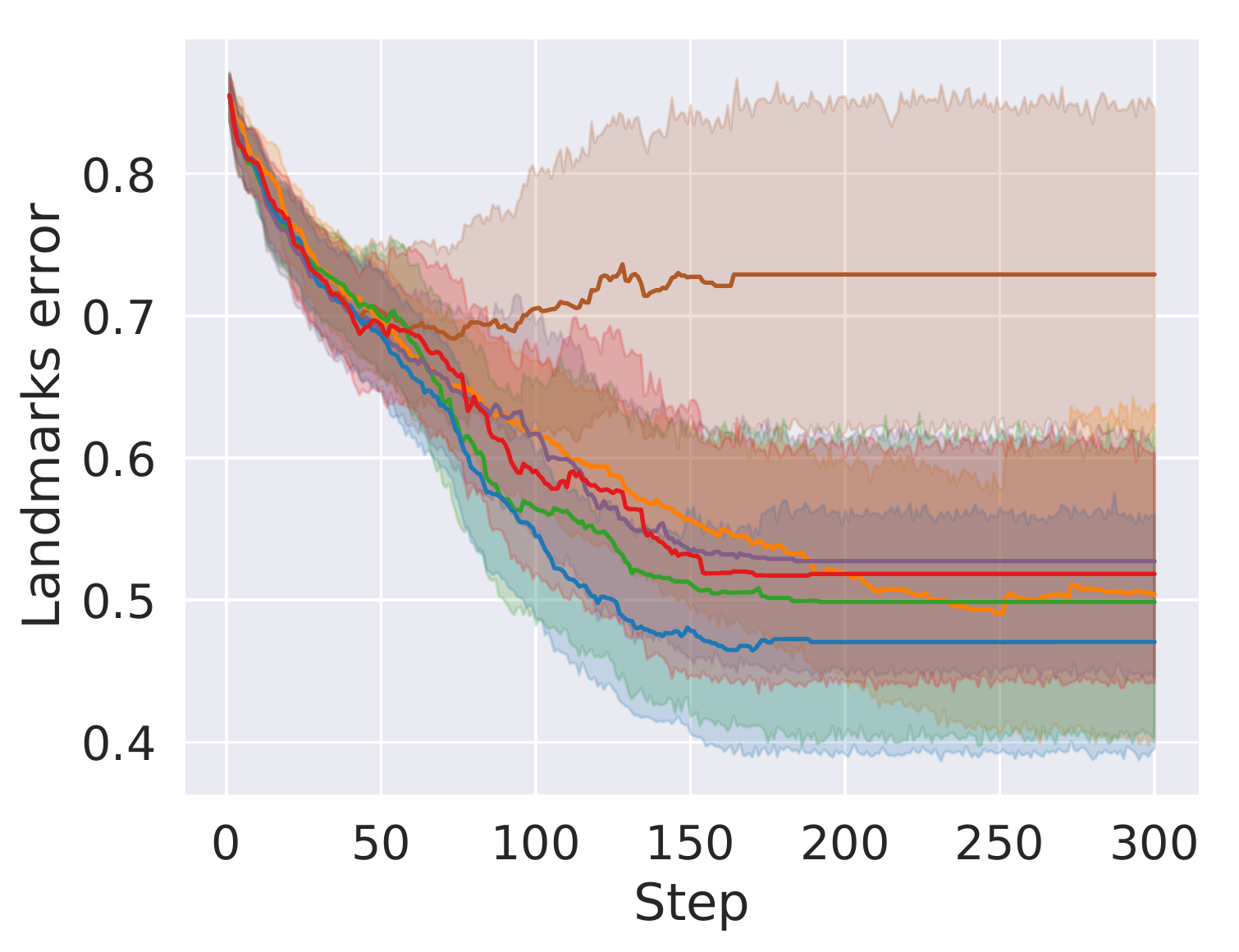}\label{40expo:landmark_error}}\
\subfigure[Max uncertainty of the trajectory]{\includegraphics[height=36mm]{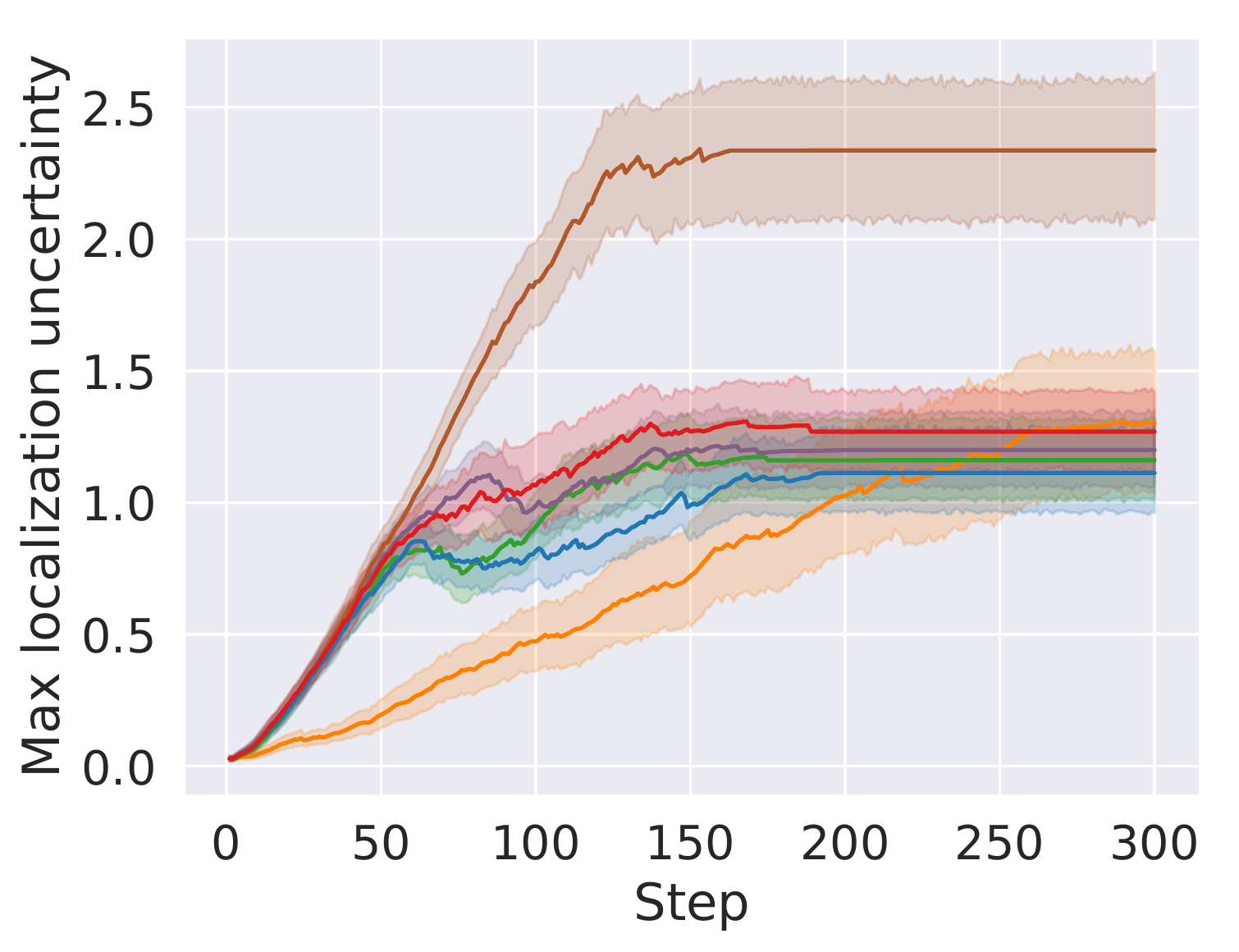}\label{40expo:loc_error}}\
\subfigure[Map entropy reduction]{\includegraphics[height=36mm]{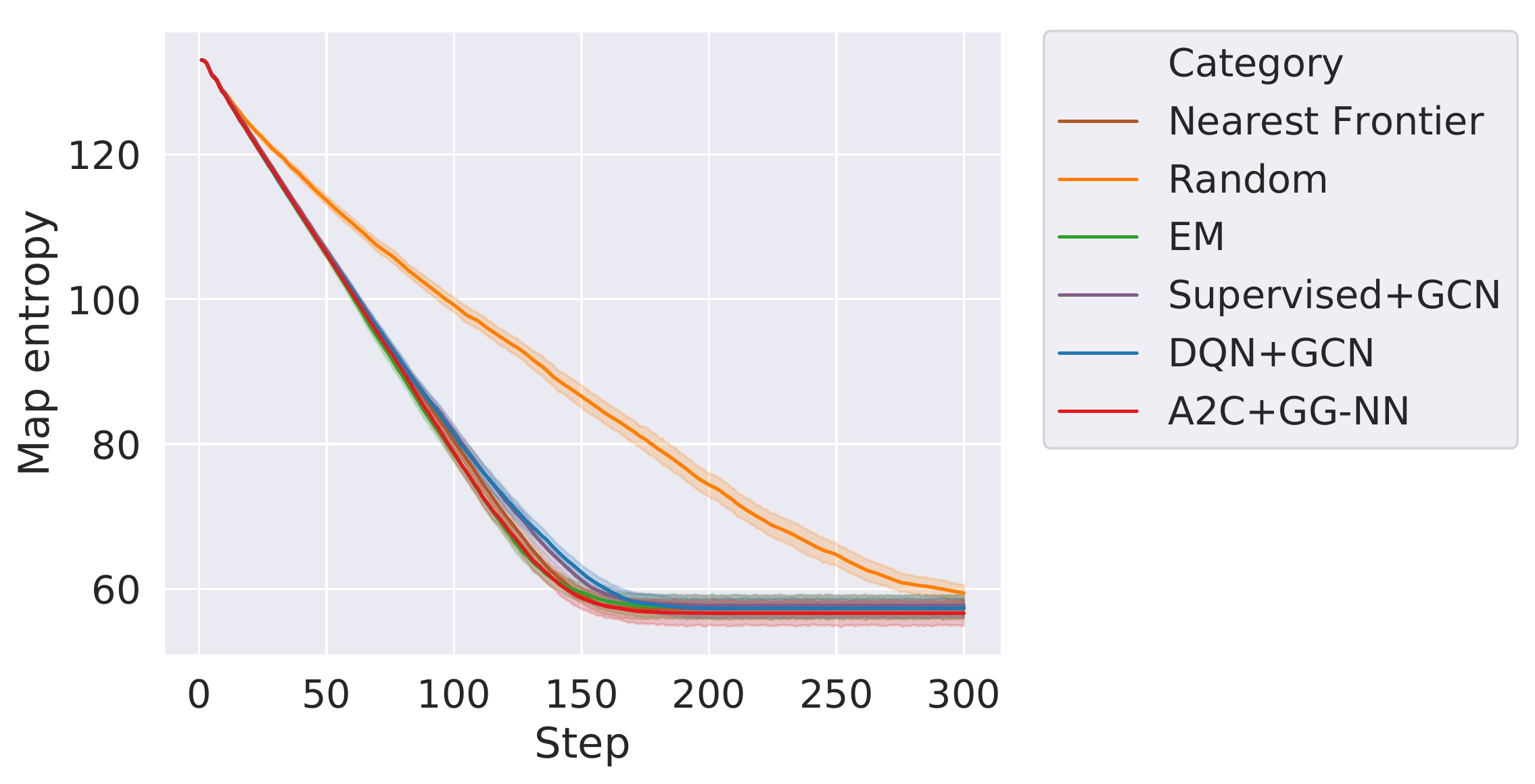}\label{40expo:map_entropy}}\
\caption{The result of 50 exploration trials of each method, with the same randomly initialized landmarks and robot start locations, on $40m \times 40m$ maps (the first three metrics shown are plotted per time-step of the simulation).}
\label{40result}
\vspace{2mm}
\end{figure*}

\begin{figure*}[ht]
\centering
\subfigure[Average uncertainty of landmarks]{\includegraphics[height=36mm]{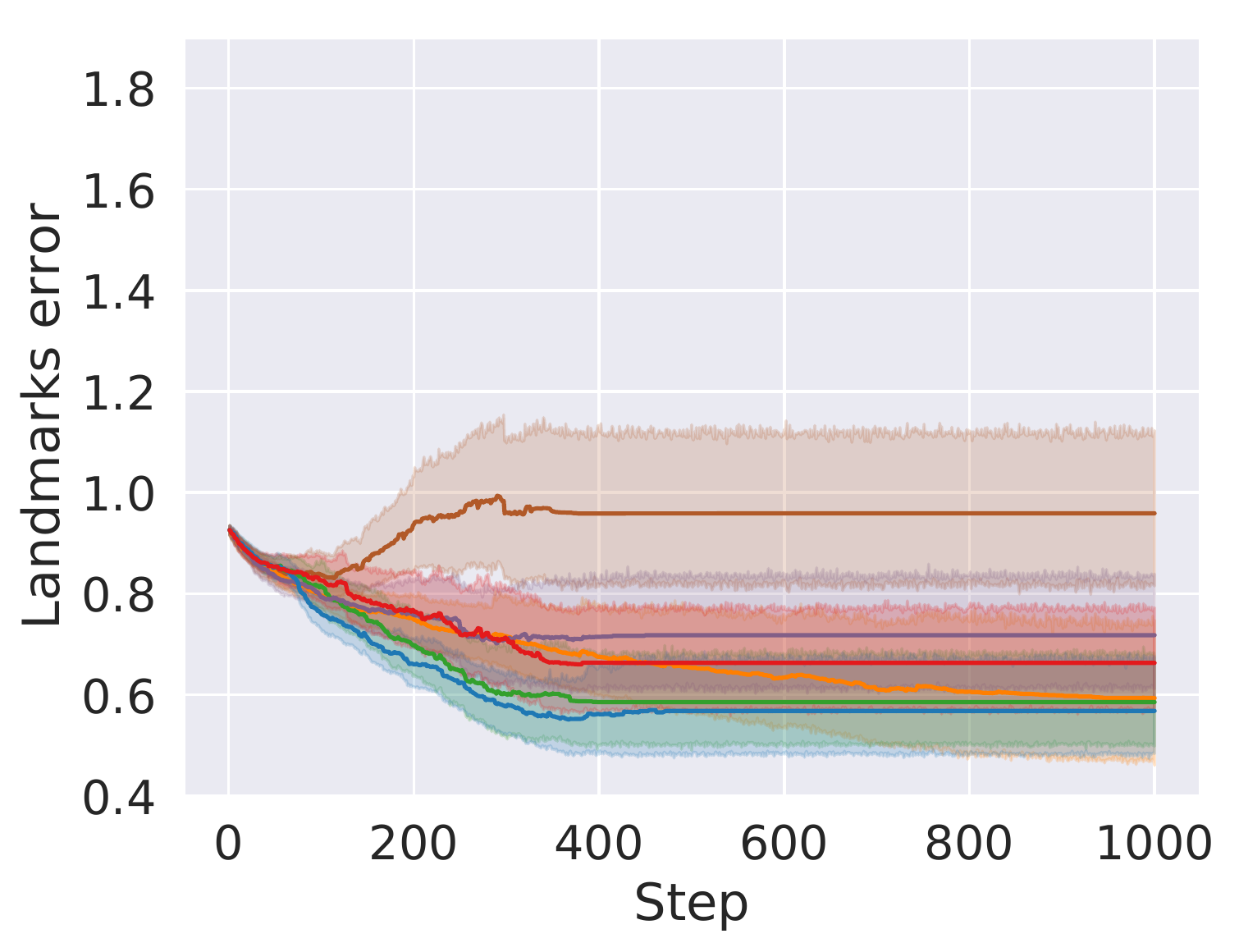}\label{60expo:landmark_error}}\
\subfigure[Max uncertainty of the trajectory]{\includegraphics[height=36mm]{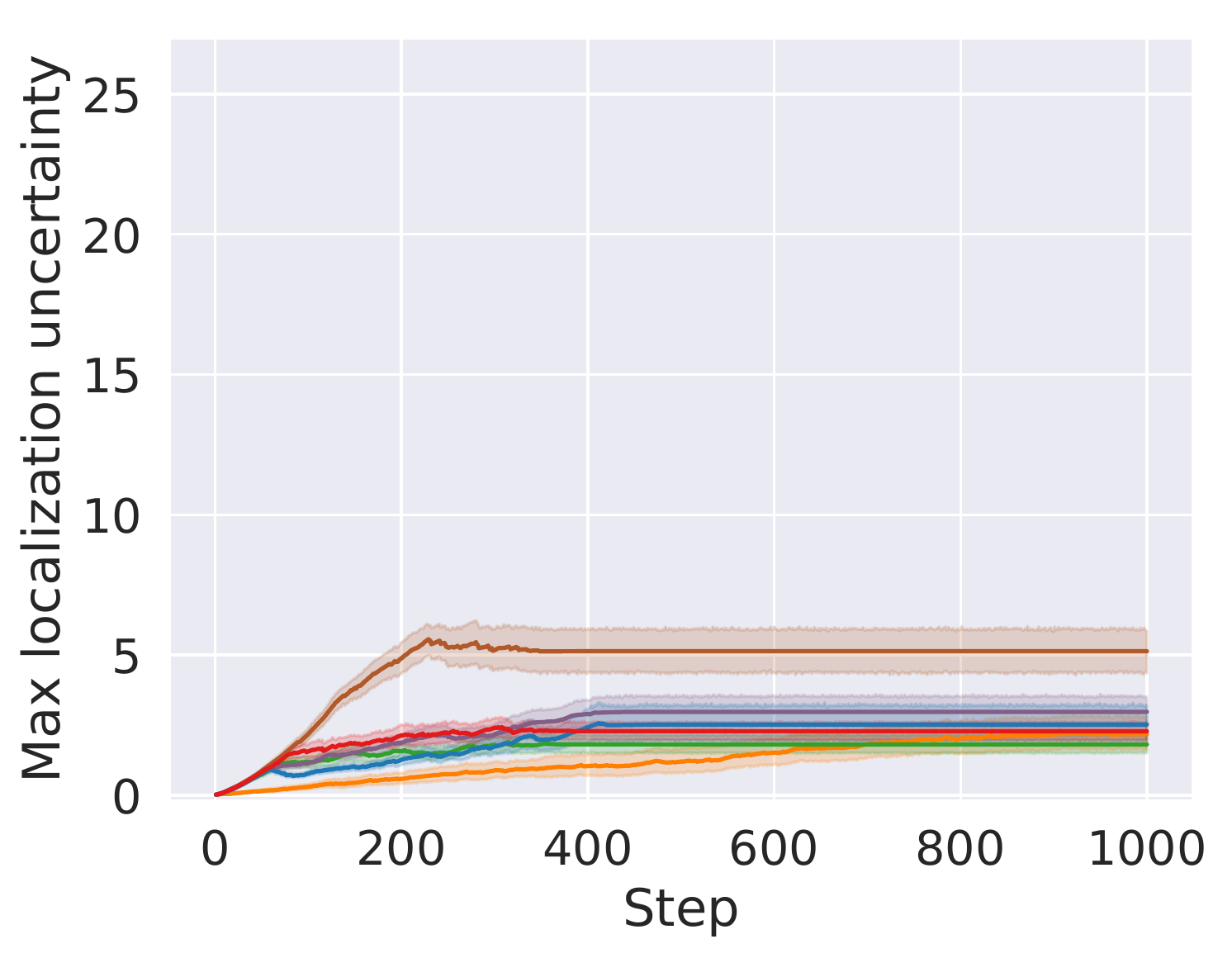}\label{60expo:loc_error}}\
\subfigure[Map entropy reduction]{\includegraphics[height=36mm]{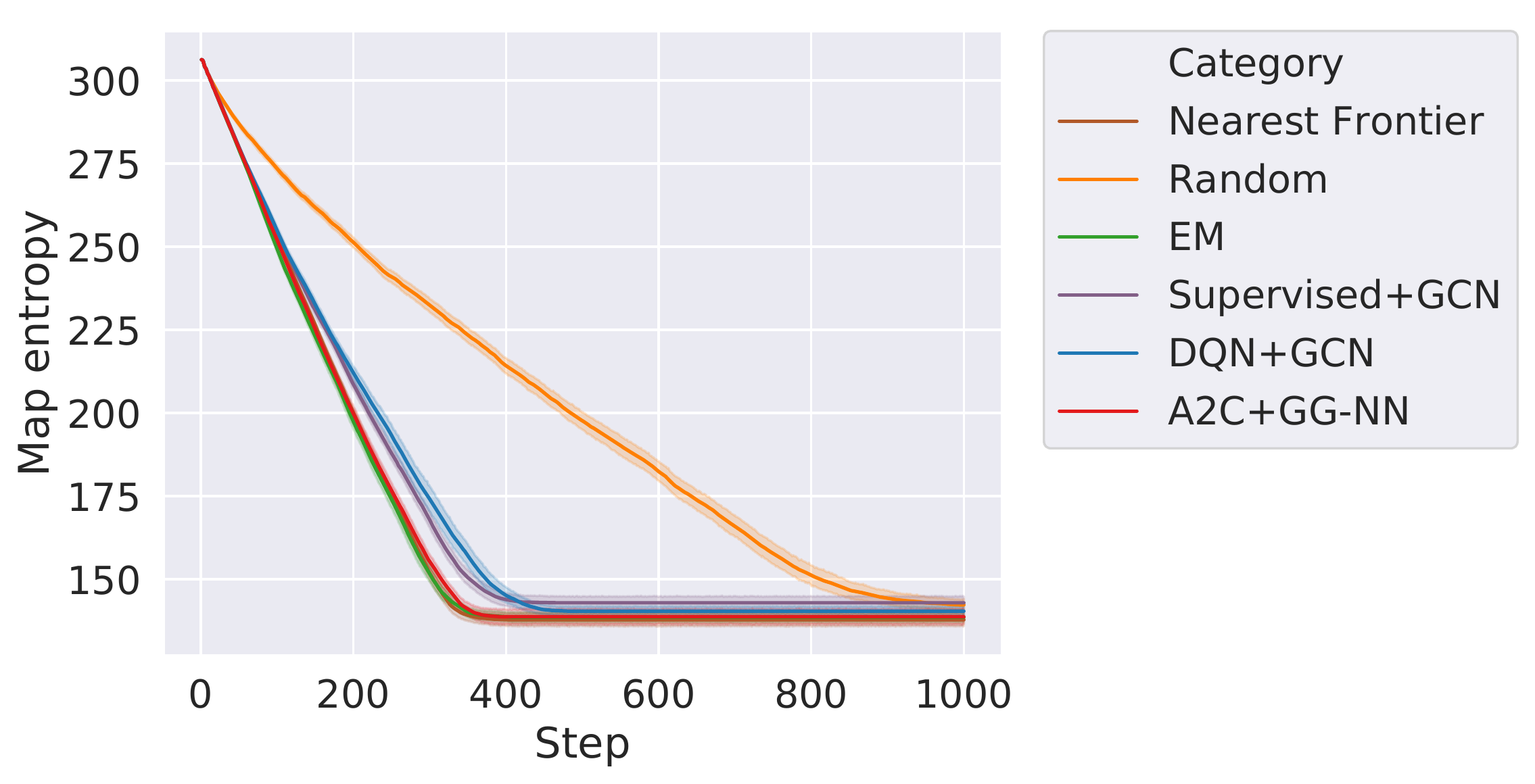}\label{60expo:map_entropy}}\
\subfigure[Average uncertainty of landmarks]{\includegraphics[height=36mm]{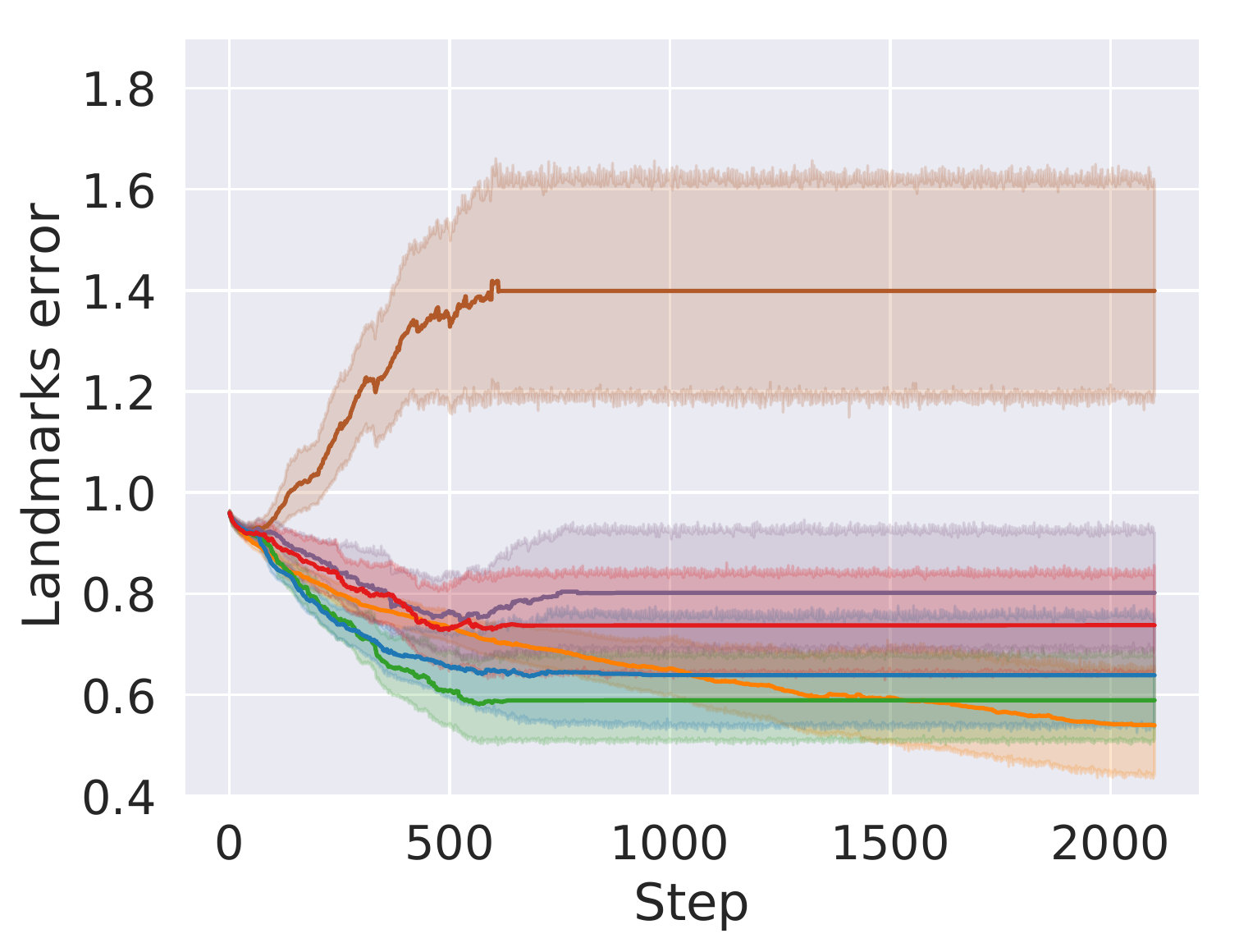}\label{80expo:landmark_error}}\
\subfigure[Max uncertainty of the trajectory]{\includegraphics[height=36mm]{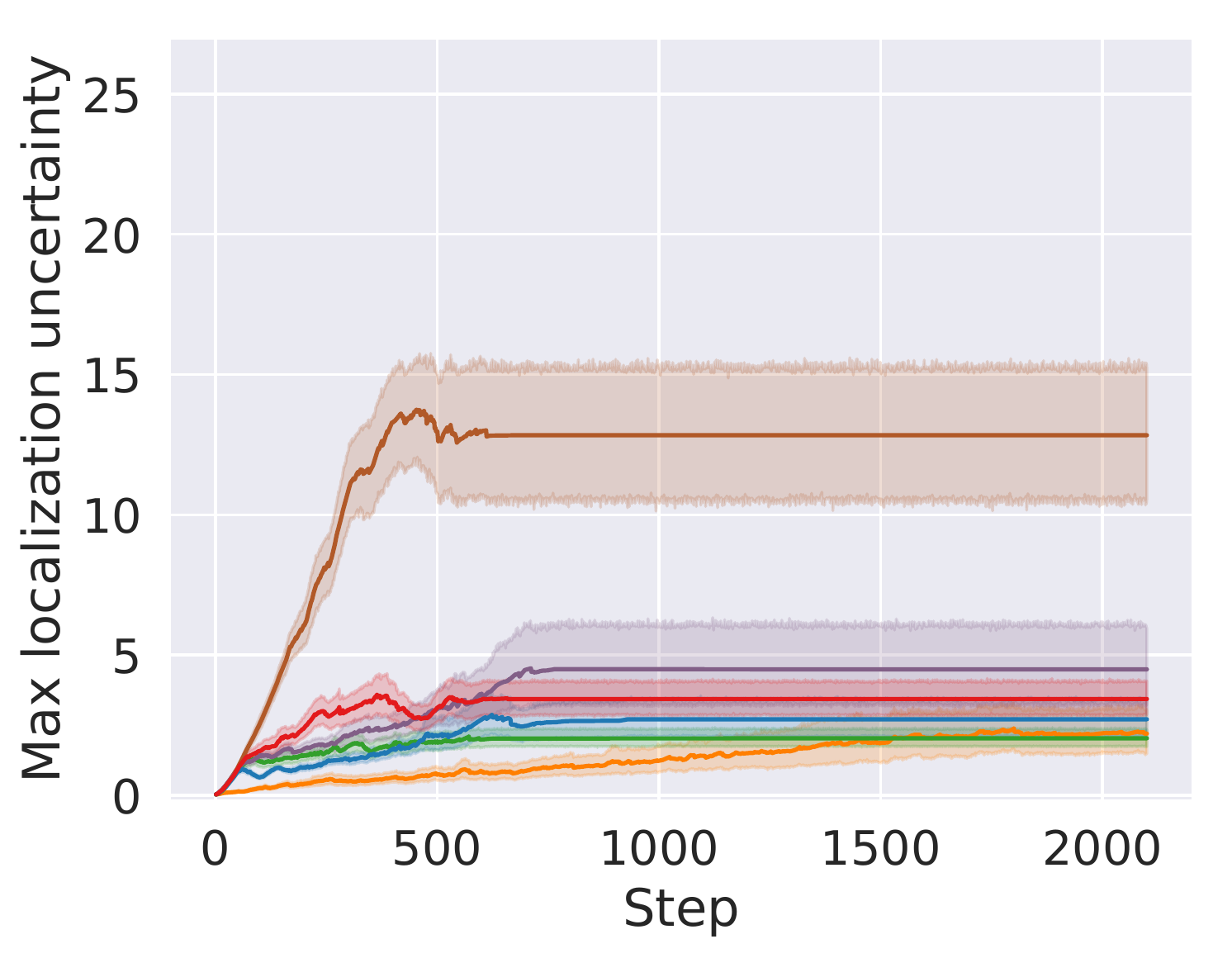}\label{80expo:loc_error}}\
\subfigure[Map entropy reduction]{\includegraphics[height=36mm]{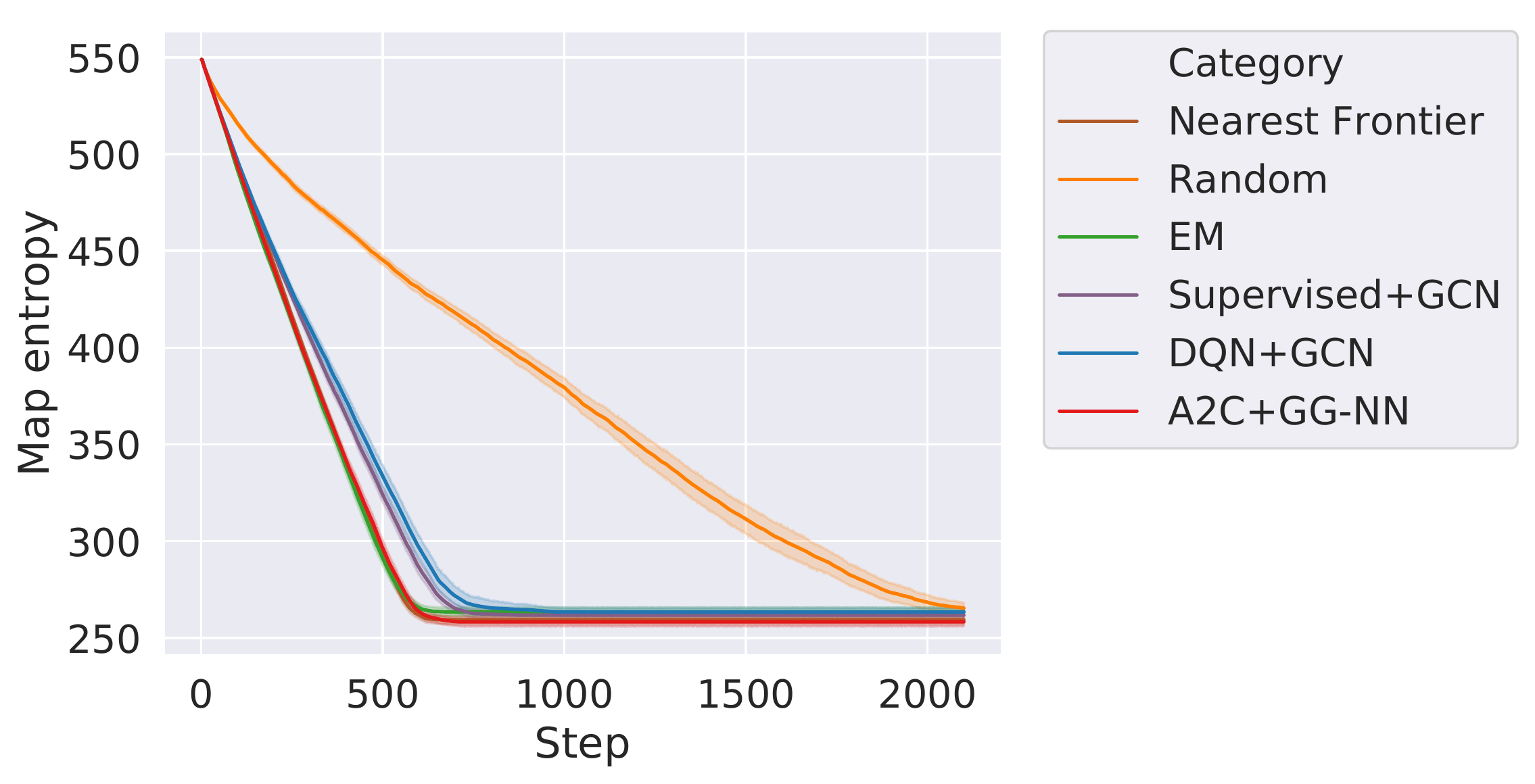}\label{80expo:map_entropy}}\
\subfigure[Average uncertainty of landmarks]{\includegraphics[height=36mm]{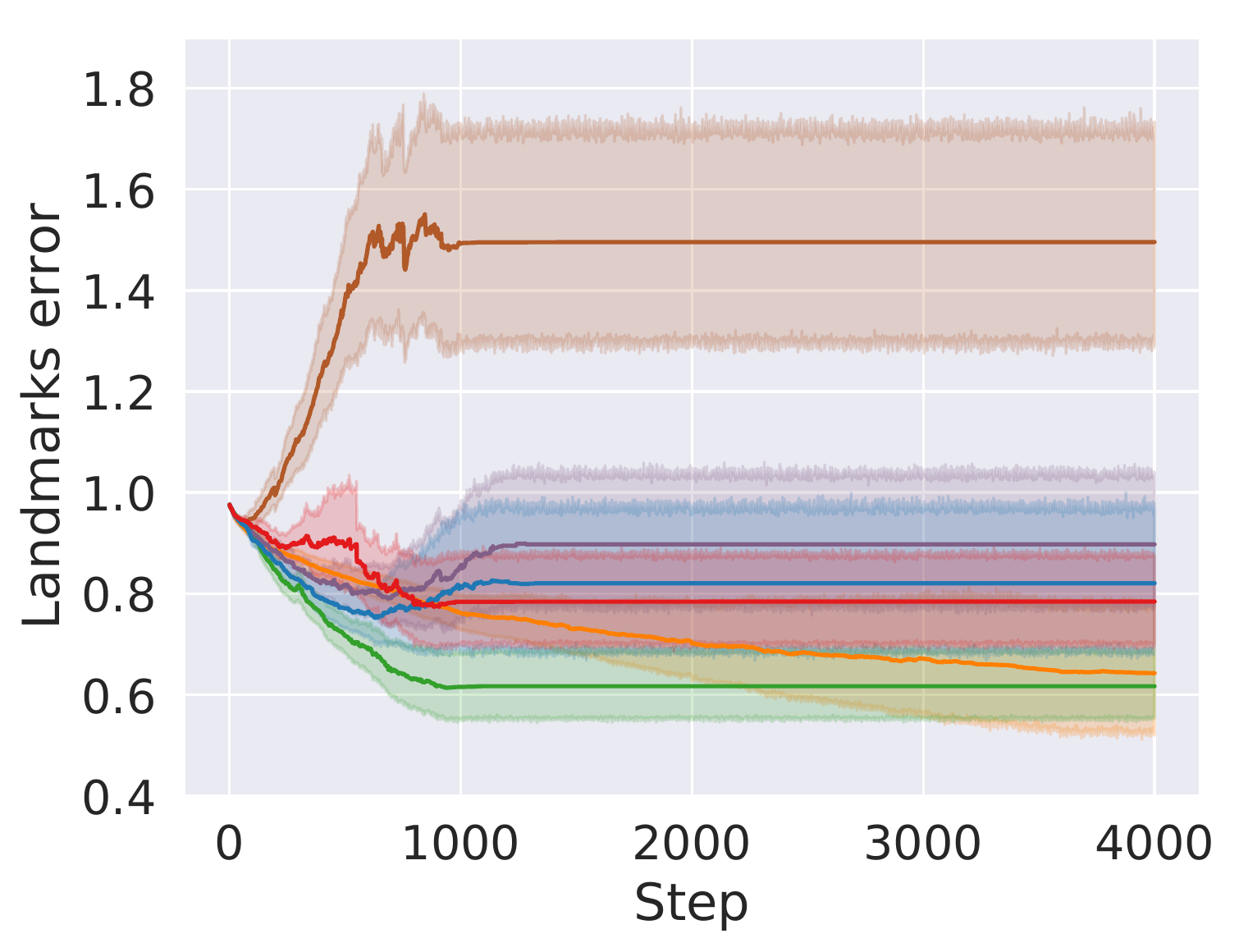}\label{100expo:landmark_error}}\
\subfigure[Max uncertainty of the trajectory]{\includegraphics[height=36mm]{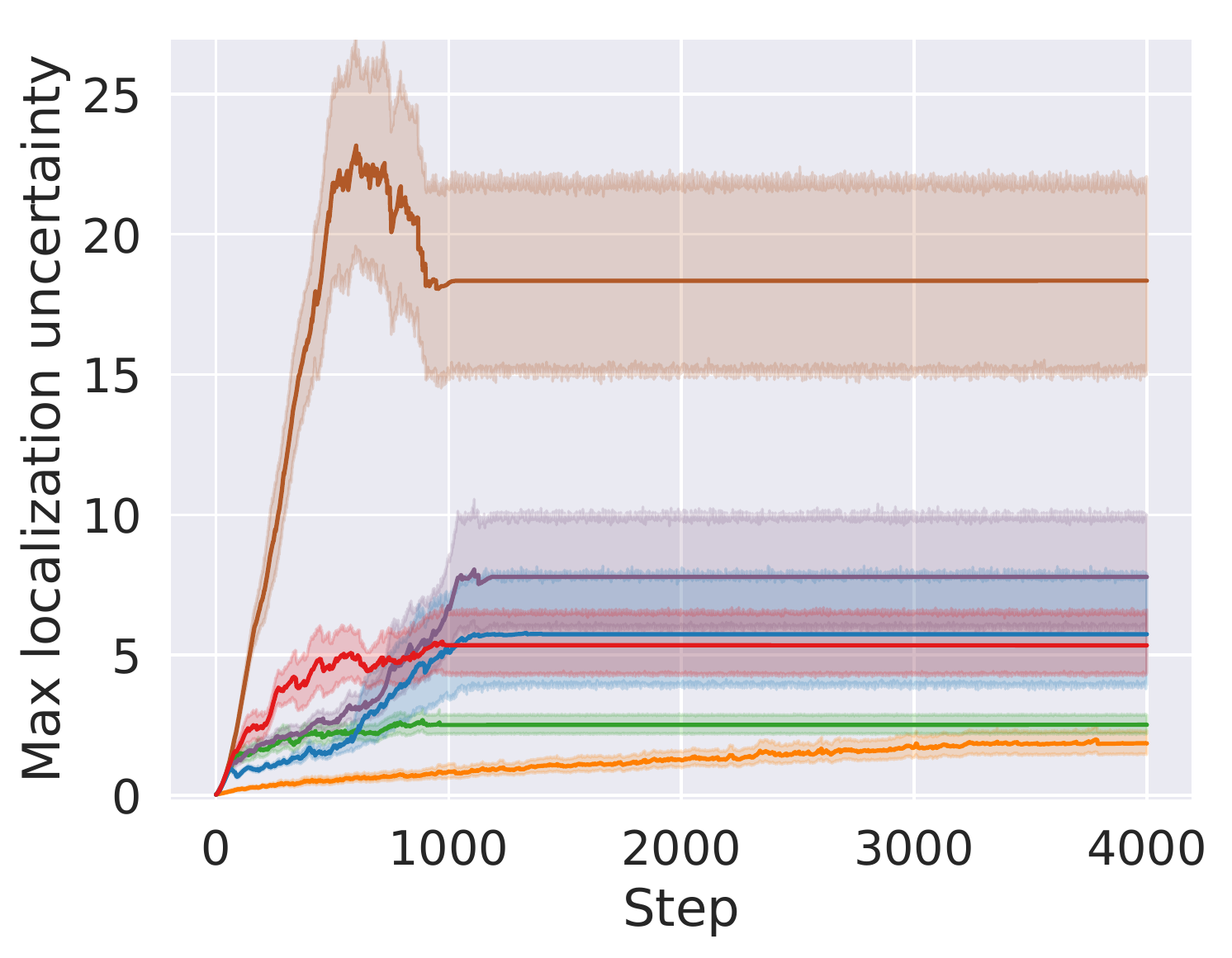}\label{100expo:loc_error}}\
\subfigure[Map entropy reduction]{\includegraphics[height=36mm]{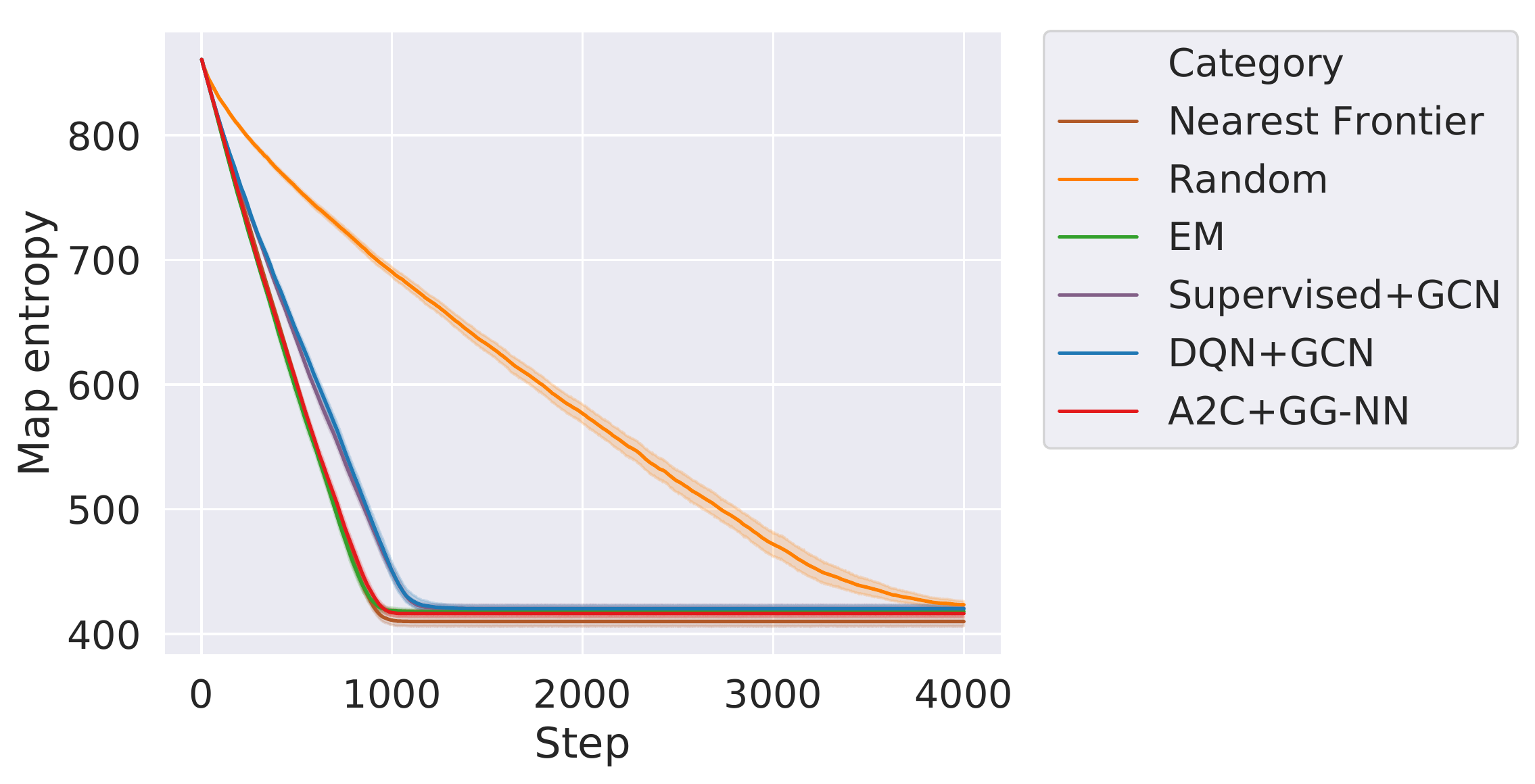}\label{100expo:map_entropy}}\
\caption{The results of 50 exploration trials on large-size maps (each of which is larger than the $40m \times 40m$ maps used for training); (a), (b), (c) represent $60m \times 60m$ maps, (d), (e), (f) represent $80m \times 80m$ maps, and (g), (h), (i) represent $100m \times 100m$ maps.}
\label{diffMap}
\end{figure*}
\subsection{Experimental Setup}

Our exploration policy is trained in a 2D simulation environment. 
The simulated robot has a horizontal field of view of $360\degree$ ($\pm 0.5\degree$). 
The measurement range of the simulated robot is $5m$ ($\pm 0.02m$). 
While exploring, the robot can rotate from $-180\degree$ to $180\degree$ ($\pm 0.2\degree$), and travel from $0m$ to $2m$ ($\pm 0.1m$) at each timestep. All simulated noise is Gaussian, and we give its standard deviation in the intervals above. 
Given the goal frontier location, the robot will first turn to the goal and then drive directly to the goal following a straight-line path. 
We use an occupancy grid map to describe the environment. 
Each occupancy map is $40m \times 40 m$ with randomly sampled landmarks and random initial robot locations. 
The \textit{feature density} of  landmarks is $0.005$ per $m^2$ in our experiment. 
To simplify the problem, landmarks are assumed passable, so obstacle avoidance is not needed.

The virtual map is made up of $2m\times 2m$ square map cells, with a $1m^2$ initial error covariance in each dimension for the virtual landmarks.
We note that virtual landmarks are only used to evaluate the reward of Alg. \ref{alg:raw_reward}, which trades map accuracy against travel expense. Meanwhile, the virtual landmarks are excluded from both SLAM factor graphs and the exploration graph. The exploration task will be terminated once $85\%$ of a map has been observed.

The simulation environment is written in Python and C++, and our graph neural network models are trained using PyTorch Geometric \cite{Fey2019}. Our code is freely available on Github\footnote{\begin{small}\url{https://github.com/RobustFieldAutonomyLab/DRL_graph_exploration}\end{small}} for use by others.

\subsection{Policy Training}
\label{subsec:policy_training}

The training environments are generated with uniformly randomly sampled landmarks and initial robot locations. A training environment example is shown in Fig. \ref{sim_env}; the uncertainty of virtual landmarks is represented by the error ellipses in each cell of the map. The blue circle represents the current field of view and the center point is the current robot location. The green error ellipses represent the uncertainty of past robot poses. The magenta points are frontier candidates, which comprise the current action space, and the red point is the selected next-view frontier. Landmarks are indicated by plus signs, which we plot in yellow once observed by the robot. Cells containing true landmarks are denoted as occupied in our map to mark their locations (ensuring a unique and non-trivial occupancy map is obtained from every exploration trial), but the landmarks are infinetesimal in size, and assumed not to present occlusions or collision hazards. 

The exploration policy is trained by DQN and A2C with three different GNN models each. The performance of each approach is shown in Fig. \ref{training_ave_reward}. For further use and study in exploration experiments, we select the policies that achieve the highest average reward for each RL framework, which are DQN+GCN and A2C+GG-NN.

\subsection{Computation Time}

The decision-making process is the most time-consuming component of exploration. The EM exploration algorithm uses forward simulation of a robot's SLAM process (including the propagation of uncertainty across virtual landmarks) to select the best next-view position from the available candidates. Hence the time complexity of the EM algorithm is $\mathcal{O}(N_{action}(C_1+C_2))$, where $N_{action}$ is the number of the candidate actions, $C_1$ is the cost of each iSAM2 predictive estimate over a candidate trajectory (a function of the number of mapped true landmarks and prior poses) and $C_2$ is the cost of the covariance update for virtual landmarks (a function of the number of prior poses and the number of virtual landmarks). We compare computation time between the EM algorithm and the RL policy on four different sizes of maps, namely $40m\times 40m$, $60m\times 60m$, $80m\times 80m$ and $100m\times 100m$. Each size has the same feature density, which is $0.005$ landmarks per $m^2$. As shown in Fig. \ref{com_time}, the average computation time of the EM algorithm grows dramatically with increasing size of the map, which leads to larger action, state and belief spaces. The EM algorithm will ultimately prove unsuitable for real-time exploration in large, complex environments. On the other hand, the RL policies use graph neural networks to predict the best action so that the computation time of the RL policy remains low, which meets the requirements for real-time application in high-dimensional state and action spaces.

\subsection{Exploration Comparison}

We compared the learned policies with (1) a nearest frontier approach, (2) the selection of a random frontier, (3) the EM approach, and (4) the GCN model trained by supervised learning in \cite{Chen2019ISRR} over 50 exploration trials. For each trial, every approach uses the same random seed to generate the same environment. Results of this comparison are shown in Figure \ref{40result}, where we evaluate three metrics: the average landmark uncertainty (of real landmarks only), the maximum localization uncertainty along the robot's trajectory, and occupancy map entropy reduction. The EM algorithm offers the best performance, but is not real-time viable in high-dimensional state-action spaces. Both the Nearest Frontier method and the Random method are real-time viable methods. However, the Nearest Frontier method has the highest landmark and localization uncertainty, and the Random method has the lowest exploration efficiency. The learning-based methods can address real-time applications and offer low landmark and localization uncertainty. The Supervised+GCN method and DQN+GCN method have similar map entropy reduction rates, but DQN+GCN offers lower landmark and localization uncertainty because it travels longer distances from one iteration to the next by selecting high-value actions. In Fig. \ref{40expo:map_entropy}, both supervised+GCN and DQN+GCN have slower entropy reduction rates than EM and the A2C+GG-NN policy.
When supervised+GCN does not perform with high accuracy, it will tend toward random action selection, because it is trained using binary labels. 

Unlike supervised learning, the RL policies are seeking the largest value for each step. 
Thus, when DQN+GCN does not have high accuracy, it will choose the highest-value action based on predicted values. Although the DQN+GCN policy has relatively low accuracy in predicting the highest-value action, it still selects high-value actions instead of random ones, which leads to more accurate mapping. The A2C+GG-NN policy has the highest exploration efficiency among learning approaches, and offers lower landmark uncertainty than the Supervised+GCN method, but not lower than the DQN+GCN method. Representative exploration trials for all of the algorithms compared in Figure \ref{40result} are provided in our video attachment.

\subsection{Scalability}

During testing, it is possible to encounter large-scale environments that induce a lengthier pose history, more landmarks, and more frontiers, generating larger exploration graphs over longer-duration operations than a robot may have been exposed to during training. We demonstrate here that our RL policies can be trained in small environments, and scale to large environments to perform exploration tasks. In this experiment, our RL policies are trained on $40m \times 40m$ maps with $0.005$ landmarks per $m^2$ feature density. The size of the testing environments are $60m \times 60m$, $80m \times 80m$ and $100m \times 100m$ with the same feature density.

The exploration results are shown in Fig. \ref{diffMap}. The A2C+GG-NN method has nearly the same exploration efficiency as the EM algorithm and the Nearest Frontier approach, which all explore faster than the Supervised+GCN and DQN+GCN methods as shown in Figs. \ref{60expo:map_entropy}, \ref{80expo:map_entropy}, and \ref{100expo:map_entropy}. The landmark uncertainty (Figs. \ref{60expo:landmark_error}, \ref{80expo:landmark_error}, \ref{100expo:landmark_error}) and  localization uncertainty (Fig. \ref{60expo:loc_error}, \ref{80expo:loc_error}, \ref{100expo:loc_error}) of our learning-based methods gradually degrade in performance with increasing map size. The Supervised+GCN method accumulates the largest landmark and localization uncertainty among learning-based methods for each map size. DQN+GCN achieves better results over $60m \times 60m$ and $80m \times 80m$ maps than A2C+GG-NN, but the performance of A2C+GG-NN is better than DQN+GCN over $100m \times 100m$ maps because of the overfitting of DQN+GCN. Overall, A2C+GG-NN demonstrates the best scalability performance based on its high exploration efficiency and relatively low landmark and localization uncertainty across all trials. Representative exploration trials showing the performance of A2C+GG-NN across all of the map sizes examined are provided in our video attachment.

\vspace{2mm}

\section{Conclusions}
\label{sec:conclusions}
We have presented a novel approach by which a mobile robot can learn, without human intervention, an effective policy for exploring an unknown environment under localization uncertainty, via exploration graphs in conjunction with GNNs and reinforcement learning. The exploration graph is a generalized topological data structure that represents the states and the actions relevant to exploration under localization uncertainty, and we build upon our previous work that used them for supervised learning with GNNs. 

Through our novel integration of this paradigm with reinforcement learning, a robot's exploration policy is shaped by the rewards it receives over time, and a designer does not need to tune hyperparameters manually, as is the case for supervised learning. Additionally, the policy learned from RL algorithms is non-myopic and robust across a variety of test environments. Our learned RL policies have exhibited the best performance among the real-time viable exploration methods examined in this paper. Moreover, we have shown that policies learned in small-scale, low-dimensional state-action spaces can be scalable to testing in larger, higher-dimensional spaces. The RL policies learned offered high exploration efficiency and relatively low map and localization uncertainty among the real-time viable exploration methods examined, across the various examples explored. In the future, we will embed obstacle information from metric maps into exploration graphs to train a robot in simulation environments and test in real-world environments.

\vspace{2mm}

\section*{Acknowledgments}

This research has been supported by the National Science Foundation, grant numbers IIS-1652064 and IIS-1723996.

\vspace{2mm}



\begin{thebibliography}{99.}%

\bibitem{Thrun2005}
S.~{Thrun}, W.~{Burgard} and D.~{Fox},
\newblock ``Exploration,''
\newblock \textit{Probabilistic Robotics}, pp. 569--605, MIT Press, 2005.

\bibitem{Julian2014}
B. J.~{Julian}, S.~{Karaman} and D.~{Rus},
\newblock ``On Mutual Information-Based Control of Range Sensing Robots for Mapping Applications,''
\newblock \textit{The International Journal of Robotics Research}, vol. 33\penalty0 (10)\penalty0, pp. 1375--1392, 2014.

\bibitem{Charrow2015CSQMI}
B.~{Charrow}, S.~{Liu}, V.~{Kumar} and N.~{Michael},
\newblock ``Information-theoretic Mapping using Cauchy-Schwarz Quadratic Mutual Information,''
\newblock \textit{Proceedings of the IEEE International Conference on Robotics and Automation}, pp. 4791--4798, 2015.

\bibitem{Chen2019ISRR}
F.~{Chen}, J.~{Wang}, T.~{Shan}, and B.~{Englot}, 
\newblock ``Autonomous Exploration Under Uncertainty via Graph Convolutional Networks,''
\newblock \textit{Proceedings of the International Symposium on Robotics Research}, 2019.



\bibitem{Jadidi2018}
M. G.~{Jadidi}, J. V.~{Mir{\'o}} and G.~{Dissanayake},
\newblock ``Gaussian processes autonomous mapping and exploration for range-sensing mobile robots,''
\newblock \textit{Autonomous Robots}, vol. 42, pp. 273--290, 2018.

\bibitem{Valencia2012}
R.~{Valencia}, J. V.~{Mir{\'o}}, G.~{Dissanayake} and J.~{Andrade-Cetto},
\newblock ``Active Pose SLAM,''
\newblock \textit{Proceedings of the IEEE/RSJ International Conference on Intelligent Robots and Systems}, pp. 1885--1891, 2012.

\bibitem{Stachniss2005}
C.~{Stachniss}, G.~{Grisetti} and W.~{Burgard},
\newblock ``Information Gain-based Exploration using Rao-Blackwellized Particle Filters,''
\newblock \textit{Proceedings of Robotics: Science and Systems}, pp. 65--72, 2005.

\bibitem{Wang2017}
J.~{Wang} and B.~{Englot},
\newblock ``Autonomous Exploration with Expectation-Maximization,''
\newblock \textit{Proceedings of the International Symposium on Robotics Research}, pp. 759--774, 2017.

\bibitem{Bai2015}
S.~{Bai}, J.~{Wang}, K.~{Doherty} and B.~{Englot},
\newblock ``Inference-Enabled Information-Theoretic Exploration of Continuous Action Spaces,''
\newblock \textit{Proceedings of the International Symposium on Robotics Research}, pp. 419--433, 2015.

\bibitem{Bai2016}
S.~{Bai}, J.~{Wang}, F.~{Chen} and B.~{Englot},
\newblock ``Information-theoretic Exploration with Bayesian Optimization,''
\newblock \textit{Proceedings of the IEEE/RSJ International Conference on Intelligent Robots and Systems}, pp. 1816--1822, 2016.

\bibitem{Bai2017}
S.~{Bai}, F.~{Chen} and B.~{Englot},
\newblock ``Toward Autonomous Mapping and Exploration for Mobile Robots through Deep Supervised Learning,''
\newblock \textit{Proceedings of the IEEE/RSJ International Conference on Intelligent Robots and Systems}, pp. 2379--2384, 2017.

\bibitem{Chen2019}
F.~{Chen}, S.~{Bai}, T.~{Shan} and B.~{Englot},
\newblock ``Self-Learning Exploration and Mapping for Mobile Robots via Deep Reinforcement Learning,''
\newblock \textit{Proceedings of the AIAA SciTech Forum}, 2019.

\bibitem{Niroui2019}
F.~{Niroui}, K.~{Zhang}, Z.~{Kashino} and G.~{Nejat},
\newblock ``Deep Reinforcement Learning Robot for Search and Rescue Applications: Exploration in Unknown Cluttered Environments,''
\newblock \textit{IEEE Robotics and Automation Letters}, vol. 4\penalty0 (2)\penalty0, pp. 610--617, 2019.

\bibitem{Scarselli2009}
F.~{Scarselli}, M.~{Gori}, A. C.~{Tsoi}, M.~{Hagenbuchner}, and G.~{Monfardini},
\newblock ``The Graph Neural Network Model,''
\newblock \textit{IEEE Transactions on Neural Networks}, vol. 20\penalty0 (1)\penalty0, pp. 61--80, 2009.

\bibitem{Sanchez2018}
A.~{Sanchez-Gonzalez}, N.~{Heess}, J. T.~{Springenberg}, J.~{Merel}, M.~{Riedmiller}, R.~{Hadsell} and P.~{Battaglia},
\newblock ``Graph Networks as Learnable Physics Engines for Inference and Control,''
\newblock \textit{Proceedings of the 35th International Conference on Machine Learning}, pp. 4470--4479, 2018.

\bibitem{Battaglia2018}
P. W.~{Battaglia}, J. B.~{Hamrick}, V.~{Bapst}, A.~{Sanchez-Gonzalez}, V.~{Zambaldi}, M.~{Malinowski}, A.~{Tacchetti}, D.~{Raposo}, A.~{Santoro}, R.~{Faulkner} and C.~{Gulcehre},
\newblock ``Relational Inductive Biases, Deep Learning, and Graph Networks,''
\newblock \textit{arXiv preprint}, arXiv:1806.01261 [cs.LG], 2018.

\bibitem{KChen2019}
K.~{Chen}, J. P.~{de Vicente}, G.~{Sepulveda}, F.~{Xia}, A.~{Soto}, M.~{Vazquez}, S.~{Savarese}
\newblock ``A Behavioral Approach to Visual Navigation with Graph Localization Networks,''
\newblock \textit{Proceedings of Robotics: Science and Systems}, 2019.

\bibitem{Wang2018}
T.~{Wang}, R.~{Liao}, J.~{Ba} and S.~{Fidler},
\newblock ``Nervenet: Learning structured policy with graph neural networks,''
\newblock \textit{Proceedings of the International Conference on Learning Representations}, 2018.

\bibitem{gtsam}
F.~{Dellaert},
\newblock ``Factor Graphs and GTSAM: A Hands-on Introduction,''
\newblock \textit{Technical Report, Georgia Institute of Technology}, GT-RIM-CP\&R-2012-002, 2012.

\bibitem{Kaess2012}
M.~{Kaess}, H.~{Johannsson}, R.~{Roberts}, V.~{Ila}, J.  J.~{Leonard} and F.~{Dellaert},
\newblock ``iSAM2: Incremental Smoothing and Mapping using the Bayes Tree,''
\newblock \textit{The International Journal of Robotics Research}, vol. 31\penalty0 (2)\penalty0, pp. 216--235, 2012.

\bibitem{Kaess2009}
M.~{Kaess} and F.~{Dellaert},
\newblock ``Covariance Recovery from a Square Root Information Matrix for Data Association,''
\newblock \textit{Robotics and Autonomous Systems}, vol. 57\penalty0 (12)\penalty0, pp. 1198-1210, 2009.

\bibitem{Kipf2017}
T. N.~{Kipf} and M.~{Welling},
\newblock ``Semi-supervised Classification with Graph Convolutional Networks,''
\newblock \textit{Proceedings of the International Conference on Learning Representations}, 2017.

\bibitem{Li2016}
Y.~{Li}, D.~{Tarlow}, M.~{Brockschmidt} and R.~{Zemel},
\newblock ``Gated Graph Sequence Neural Networks,''
\newblock \textit{Proceedings of the International Conference on Learning Representations}, 2016.

\bibitem{Gao2019}
H.~{Gao} and S.~{Ji},
\newblock ``Graph U-Nets,''
\newblock \textit{Proceedings of the International Conference on Learning Representations}, 2019.

\bibitem{Ronneberger2015}
O.~{Ronneberger}, P.~{Fischer} and T.~{Brox},
\newblock ``U-net: Convolutional networks for biomedical image segmentation,''
\newblock \textit{International Conference on Medical Image Computing and Computer-assisted Intervention}, pp. 234-241, 2015.

\bibitem{Puterman1994}
M. L.~{Puterman},
\newblock ``Model Formulation,''
\newblock \textit{Markov Decision Processes: Discrete Stochastic Dynamic Programming}, pp. 17--32, John Wiley \& Sons, 1994.

\bibitem{Gal2015}
Y.~{Gal} and Z.~{Ghahramani},
\newblock ``Dropout as a Bayesian approximation: Representing model uncertainty in deep learning,''
\newblock \textit{Proceedings of the International Conference on Machine Learning}, pp. 1050-1059, 2016.

\bibitem{Mnih2015}
V.~{Mnih}, K.~{Kavukcuoglu}, D.~{Silver}, A. A.~{Rusu}, J.~{Veness}, M. G.~{Bellemare}, A.~{Graves}, M.~{Riedmiller}, A. K.~{Fidjeland}, G.~{Ostrovski} and S.~{Petersen},
\newblock ``Human-level Control through Deep Reinforcement Learning,''
\newblock \textit{Nature}, vol. 518\penalty0 (7540)\penalty0, pp. 529--533, 2015.

\bibitem{Vinyals2017}
O.~{Vinyals}, T.~{Ewalds}, S.~{Bartunov}, P.~{Georgiev}, A. S.~{Vezhnevets}, M.~{Yeo}, A.~{Makhzani}, H.~{Küttler}, J.~{Agapiou}, J.~{Schrittwieser}, J.~{Quan}, S.~{Gaffney}, S.~{Petersen}, K.~{Simonyan}, T.~{Schaul}, H. v.~{Hasselt}, D.~{Silver}, T.~{Lillicrap}, K.~{Calderone}, P.~{Keet}, A.~{Brunasso}, D.~{Lawrence}, A.~{Ekermo}, J.~{Repp} and R.~{Tsing},
\newblock ``Starcraft II: A New Challenge for Reinforcement Learning,''
\newblock \textit{arXiv preprint}, arXiv:1708.04782 [cs.LG], 2017.

\bibitem{Fey2019}
M.~{Fey} and J. E.~{Lenssen},
\newblock ``Fast Graph Representation Learning with PyTorch Geometric,''
\newblock \textit{ICLR Workshop on Representation Learning on Graphs and Manifolds}, 2019.

\end{thebibliography}
\end{document}